\PassOptionsToPackage{square,sort,comma,numbers}{natbib}
\PassOptionsToPackage{table}{xcolor}
\documentclass{article}

\usepackage[preprint]{neurips_2026}

\usepackage[utf8]{inputenc} % allow utf-8 input
\usepackage[T1]{fontenc}    % use 8-bit T1 fonts
\usepackage{hyperref}       % hyperlinks
\usepackage{url}            % simple URL typesetting
\usepackage{booktabs}       % professional-quality tables
\usepackage{amsfonts}       % blackboard math symbols
\usepackage{nicefrac}       % compact symbols for 1/2, etc.
\usepackage{microtype}      % microtypography
\usepackage{xcolor}         % colors
\usepackage{pifont}
\usepackage{times}
\usepackage{latexsym}

\usepackage[T1]{fontenc}
\usepackage[utf8]{inputenc}

\usepackage{microtype}
\usepackage{comment}
\usepackage{inconsolata}

\usepackage{multirow}
\usepackage{graphicx}
\usepackage{xcolor}
\usepackage{tcolorbox}
\usepackage{fontawesome5}
\usepackage{xcolor} % For text color
\usepackage{enumitem} 
\usepackage{tabularx}
\usepackage{booktabs}
\usepackage{graphicx}
\usepackage{arydshln}
\usepackage{amssymb}
\usepackage{xcolor}
\usepackage{tabularx, booktabs, xcolor, colortbl}
\usepackage{graphicx}
\usepackage{multirow}
\usepackage{tabularray}
\usepackage{makecell}
\usepackage{ragged2e}
\newcolumntype{Y}{>{\RaggedRight\arraybackslash}X}

\definecolor{Gray}{gray}{0.9}
\definecolor{keywordcolor}{RGB}{0,0,139}
\definecolor{variablecolor}{RGB}{0,100,0}
\definecolor{green}{RGB}{0,255,0}
\definecolor{lightgreen}{RGB}{179, 247, 96}
\definecolor{blue}{RGB}{0,0,255}
\definecolor{orange}{RGB}{255,165,0}
\definecolor{red}{RGB}{255,0,0}
\definecolor{purple}{RGB}{128,0,128}
\definecolor{cyan}{RGB}{0,255,255}
\definecolor{magenta}{RGB}{255,0,255}
\definecolor{yellow}{RGB}{255,255,0}
\definecolor{brown}{RGB}{139,69,19}
\definecolor{gray}{RGB}{128,128,128}
\definecolor{pink}{RGB}{255,182,193}
\definecolor{teal}{RGB}{0,128,128}
\definecolor{olive}{RGB}{128,128,0}
\definecolor{lightblue}{RGB}{173,216,230}
\definecolor{darkblue}{RGB}{0,0,139}

\definecolor{lightgreen}{rgb}{0.56, 0.93, 0.56}
\definecolor{lightyellow}{rgb}{0.96, 0.93, 0.26}
\definecolor{customblue}{rgb}{0.2, 0.6, 1.0}
\definecolor{pink}{rgb}{1.0, 0.6, 0.8}

\usepackage{listings}
\usepackage{xcolor}
\usepackage{tcolorbox}
\usepackage{amsmath}

\definecolor{commentgreen}{rgb}{0,0.5,0}    % Green comments
\definecolor{keywordblue}{rgb}{0,0,1}       % Blue keywords
\definecolor{stringred}{rgb}{0.6,0,0}       % Red strings
\definecolor{numberpurple}{rgb}{0.58,0,0.82} % Purple numbers

\lstdefinestyle{custompython}{
    commentstyle=\color{commentgreen},  % Green comments
    keywordstyle=\bfseries\color{keywordblue}, % Bold blue keywords
    stringstyle=\color{stringred},  % Red strings
    numberstyle=\tiny\color{numberpurple},  % Purple numbers
    basicstyle=\ttfamily\footnotesize, % Monospace font
    numbers=left,    % Line numbers on the left
    numbersep=5pt,   % Space between numbers and code
    frame=single,    % Frame around the code
    framerule=0.5pt, % Thickness of frame
    tabsize=4,       % Tab size
    breaklines=true, % Automatic line breaking
    captionpos=b,    % Caption position (bottom)
}

\newtcolorbox[auto counter, number within=section]{mybox}[2][]{colframe=codeblue!80!black, colback=backcolour!90, coltitle=white, fonttitle=\bfseries, title=Example~\thetcbcounter: #2,#1}

\title{ReasonLab: A Controlled and Auditable Evaluation of\\Prompting Techniques for Multiple-Choice QA}
\vspace{-5mm}

\author{%
  Inder Preet \thanks{Corresponding authors}\\
  IBM, Dublin \\
  \texttt{inder.preet@ibm.com} \\
  \And
  Shuxin Lin \\
  IBM, New York\\
  \texttt{shuxin.lin@ibm.com} \\
  \And 
  Dhaval Patel\footnotemark[1] \\
  IBM, New York \\
  \texttt{pateldha@us.ibm.com} \\
}

\begin{document}
% Global spacing adjustments
% \setlength{\parskip}{-0.5mm}
\setlength{\abovecaptionskip}{2pt}
\setlength{\belowcaptionskip} {0pt}
\maketitle
\vspace{-3mm}
\begin{abstract}
Probing the capabilities of Large Language Models (LLMs) and building robust solutions
for Multiple-Choice Question Answering (MCQA) remain central challenges in natural
language understanding. Furthermore, the rapid proliferation of LLMs has created the
implicit assumption that more sophisticated prompting techniques yield better
performance. Several studies claim better performance with more sophisticated prompting
techniques, but report them under differing models, prompt wordings, decoding
configurations and answer-extraction rules, so that the reported gains cannot be
attributed to the technique alone. We address this gap with \textbf{ReasonLab}, an
evaluation framework in which the prompting technique is a first-class experimental
variable alongside the model and the dataset, and which retains every individual
generation together with its token counts, whether it ran to completion or was cut off at
the output limit, and how its answer was extracted. Using ReasonLab we conduct a controlled empirical study of
8~prompting techniques across 10~MCQA datasets, encompassing 27~model configurations and
roughly 4{,}300 unique questions evaluated 480{,}927 times in total, where each
evaluation is a unique question by configuration pair scored from a single generation at
\texttt{temperature=0}. Our findings reveal that the choice of prompting technique is a
minor determinant of accuracy because on the configurations evaluated without a reasoning budget
the reasoning-trigger techniques improve on direct prompting by only 3.92 to 4.69~pp and
are statistically indistinguishable from one another, and on the configurations evaluated
with reasoning enabled no technique differs from direct prompting by more than 0.51~pp,
with Self-Generate the only technique showing a consistent effect, namely a reduction of
2.95~pp. We further investigate three critical phenomena: (1) the comparison of models
on a common set of datasets, where model size does not predict accuracy, (2) the performance-efficiency trade-offs across model
variants with different thinking budgets, revealing that enabling reasoning is worth up to
12.74~pp whereas an eightfold budget increase adds only 0.48 to 2.10~pp, and (3) the substantial variation in dataset difficulty, with
60\% of benchmarks below 70\% accuracy and a 43.9~pp spread from
easiest to hardest, indicating considerable room for model improvement. These results
suggest that, for MCQA, the prompting technique is a minor lever compared with enabling
model reasoning, and that substantial performance gaps remain across diverse benchmarks,
offering opportunities for genuine model improvements rather than prompt optimization.
\end{abstract}
\vspace{-3mm}
\section{Introduction}
\vspace{-3mm}
The field of LLM evaluation has witnessed an explosion of prompting techniques, each
claiming to unlock better reasoning. From Chain-of-Thought
(CoT)~\cite{wei2022chain} to self-consistency~\cite{wang2023selfconsistencyimproveschainthought}
and a growing family of plan-and-solve, analogical and self-generated variants, the
community has invested considerable effort in developing increasingly sophisticated
prompting strategies. This has fostered an implicit assumption that complex prompting is
necessary for strong performance. That assumption is difficult to test from the published
record, because each technique is reported under its own model set, prompt wording,
decoding configuration and answer-extraction rule, and thus the reported gains cannot be
attributed to the technique alone.

We introduce \textbf{ReasonLab}, an evaluation framework designed for that comparison.
Existing harnesses hold the prompt fixed and vary the model. ReasonLab instead treats the
prompting technique as a first-class experimental variable, expressing each technique as a
substitution at one of five interchangeable points in the MCQA pipeline
(\S\ref{sec:methodology}), i.e., question presentation, context construction, reasoning
trigger, answer generation and answer extraction. As such, any technique in the literature
can be run against any model and dataset under a single protocol. ReasonLab additionally
retains, for every (technique, model, dataset, question) cell, the generated text, the
prompt and completion token counts, whether the generation stopped on its own or was
truncated at the output limit, and how the answer was extracted from it. Accuracy is
therefore reported jointly with cost, and every comparison can be traced back to the
individual generations that produced it.

Using ReasonLab we carried out 480{,}927 model-question evaluations spanning
8~prompting techniques, 10~MCQA benchmarks and 27~model-configuration pairs, covering
13~distinct models with multiple reasoning budgets where applicable. Each evaluation is a
unique question by configuration pair scored from a single generation at
\texttt{temperature=0}, i.e., the count reflects coverage rather than repeated sampling of
the same item. The approximately 4{,}300 unique questions provide the power for the paired
cell-level tests we report.

We focus on MCQA because it remains the dominant evaluation primitive in the LLM
literature and underlies most public leaderboards, while also being the setting in which
prompting techniques have been most aggressively promoted. Although this scoping limits
direct claims about open-ended or agentic settings, it enables a controlled and
well-powered comparison that prior empirical studies of prompting have not provided at
this scale.

\subsection{Key Findings}
\vspace{-2mm}
Our analysis yields three findings. Each is stated for the regime in which it was
measured, i.e., multiple-choice question answering, a single generation per question at
\texttt{temperature=0}, the model configurations listed for that comparison, and
techniques implemented from their published templates without per-model tuning.

\textbf{1. The choice of prompting technique has a small effect on accuracy, and that
effect does not distinguish one technique from another.} On the 8 configurations
evaluated without a reasoning budget (\S\ref{sec:matched}), five techniques improve on
direct prompting, i.e., Self-Analogical~\cite{yasunaga2024large} by 4.69~pp,
Plan-and-Solve~\cite{wang2023planandsolve} by 4.61~pp, CoT~\cite{wei2022chain} by
4.41~pp, and the CoT-Expert and CoT-Inductive role primers by 4.25~pp and 3.92~pp, each
significant at $p<10^{-6}$. However, the differences among these five are not significant. On the 4 open-weight configurations
evaluated with reasoning enabled (GPT-OSS~\cite{gptoss2025} and Qwen3~\cite{qwen3};
\S\ref{sec:armb}), no technique differs from direct prompting by more than 0.51~pp, and none of the differences is significant ($p\geq0.17$,
paired $t$-test; $p\geq0.19$, Wilcoxon signed-rank). Self-Generate~\cite{yu2022generate} is the only
technique with a consistent effect across both regimes, and that effect is a reduction in
accuracy of 2.95~pp ($p<10^{-4}$). Prompt-Repeat~\cite{promptrepeat} is the remaining
technique and it does not improve on direct prompting in either regime.

\textbf{2. Enabling reasoning is the dominant lever, whereas scaling the reasoning
budget is not.} We control reasoning effort through a token budget added to the
generation limit, i.e., \emph{low}~=~1{,}024, \emph{medium}~=~8{,}192 and
\emph{high}~=~24{,}576 reasoning tokens on top of a 4{,}096-token content limit. Where
the no-reasoning setting is verified in the logs to emit zero reasoning tokens, enabling
reasoning adds \textbf{+12.74~pp} for GPT-5.4~\cite{gpt54card} and \textbf{+2.10~pp} for
Claude Sonnet~4~\cite{claude4card}, i.e., the same intervention is worth six times as much
on one model as on the other.
Within the enabled state, an eightfold increase in budget from \emph{low} to \emph{medium}
adds only 0.48 to 2.10~pp while consuming a further 342 to 788 reasoning tokens per
question, and at \emph{medium} the models spend between 7\% and 20\% of the budget they
are given. GPT-5~\cite{gpt5card} (\texttt{2025-08-07}) still reasons on 18.2\% of questions
when no reasoning effort is requested, and Gemini~2.5~Pro~\cite{gemini25} offers no setting
that suppresses reasoning, so both contribute to the budget-scaling comparison only.

\textbf{3. The benchmarks in our suite are not saturated.} 6 of the 10
datasets sit below 70\% mean accuracy across all evaluated configurations, and 4 of 10
remain below 70\% even for the best configuration on each dataset. The spread from the
easiest to the hardest dataset is 43.9~pp. If some of these questions
already appear in the training data of the models we evaluate, that would push the
measured scores up rather than down, so the headroom we report is more likely an
underestimate than an overestimate (\S\ref{sec:difficulty}).

\section{Related Work}
\vspace{-3mm}
\label{sec:relatedwork}
\textbf{Prompting Techniques for LLMs}
The prompting literature has grown rapidly since Chain-of-Thought (CoT)
prompting~\cite{wei2022chain} demonstrated that eliciting
step-by-step reasoning can improve performance on arithmetic and commonsense
tasks, spawning Zero-shot CoT~\cite{kojima2022large},
Self-Consistency~\cite{wang2023selfconsistencyimproveschainthought},
Plan-and-Solve~\cite{wang2023planandsolve},
Self-Ask~\cite{press2023measuring},
Tree-of-Thoughts~\cite{yao2023tree}, and analogical
prompting~\cite{yasunaga2024large}. The recent Prompt
Report~\cite{schulhoff2024prompt} catalogues over 50 distinct techniques. A
parallel critical strand has questioned how robust the reported gains
actually are: prompt formatting can swing accuracy by tens of
points~\cite{sclar2024quantifying, mizrahi2024state, lu2022fantastically},
CoT primarily helps on math and symbolic tasks~\cite{sprague2024cot}, and
self-verification frequently fails to improve answers~\cite{stechly2024self}.

Sprague et al.~\cite{sprague2024cot} is the closest prior work and reaches a compatible
conclusion, namely that CoT helps mainly on mathematical and symbolic tasks, by a
different route. Theirs is a meta-analysis over published results, and as such it
inherits each source paper's model, prompt wording, decoding configuration and answer
extractor, and can compare only the numbers those pipelines reported. Ours is a
controlled primary study in which one parser, one protocol and one serving configuration
are held fixed while the technique varies, so that every comparison is paired on
identical questions and identical models. We also differ in coverage, i.e., 8 techniques
rather than CoT against direct prompting, together with a decomposition of the reasoning
budget and token-cost accounting.

\textbf{Evaluation Frameworks and Benchmarks}
LLM evaluation infrastructure has matured along two axes. Broad benchmark suites such as HELM~\cite{liang2022holistic}, BIG-bench~\cite{srivastava2022beyond}, MMLU~\cite{hendrycks2020measuring}, and the lm-evaluation-harness~\cite{eval-harness} provide wide task coverage and standardized scoring, but treat the prompt as fixed and focus on cross-model comparison. Closer to our setting, PromptBench~\cite{zhu2024promptbench} and the Prompt Report~\cite{schulhoff2024prompt} systematically vary prompts, but neither pairs technique-level comparison with reasoning-budget analysis under a unified protocol.

Table~\ref{tab:framework_comparison} situates ReasonLab in this landscape: an evaluation harness, not a benchmark, that additionally logs reasoning and output-token counts so accuracy can be reported jointly with cost.
\begin{table}[h]
\vspace{-3mm}
\centering
\caption{Comparison with related evaluation infrastructure. ``Multi-Technique''
indicates systematic variation of prompting strategy; ``Token Analysis'' indicates
per-question reasoning/output-token logging; ``Cell Provenance'' indicates that every individual
generation is retained together with its token counts, whether it was truncated, and how
its answer was extracted, so that any reported aggregate can be traced back to the
generations behind it.}
\label{tab:framework_comparison}
\small
\setlength{\tabcolsep}{4pt}
\begin{tabular}{lcccc}
\toprule
\textbf{Framework} & \textbf{Multi-Technique} & \textbf{Multi-Dataset} & \textbf{Token Analysis} & \textbf{Cell Provenance} \\
\midrule
HELM~\cite{liang2022holistic} & \ding{55} & \ding{51} & \ding{55} & Partial \\
BIG-bench~\cite{srivastava2022beyond} & \ding{55} & \ding{51} & \ding{55} & \ding{55} \\
lm-eval-harness~\cite{eval-harness} & \ding{55} & \ding{51} & \ding{55} & Partial \\
PromptBench~\cite{zhu2024promptbench} & \ding{51} & \ding{51} & \ding{55} & \ding{55} \\
Prompt Report~\cite{schulhoff2024prompt} & \ding{51} & Partial & \ding{55} & \ding{55} \\
\textbf{ReasonLab (Ours)} & \ding{51} & \ding{51} & \ding{51} & \ding{51} \\
\bottomrule
\end{tabular}
\end{table}
\vspace{-3mm}

\textbf{Reasoning Models and Test-Time Compute}
Recent reasoning-tuned models such as OpenAI's o1/o3 family, GPT-5, Claude with extended thinking, DeepSeek-R1~\cite{deepseekai2025r1}, and Qwen3~\cite{qwen3} expose a configurable thinking budget at inference time. A growing body of work examines how additional test-time compute trades off against accuracy~\cite{snell2024scaling, muennighoff2025s1}, and finds that returns are often non-monotonic and model-dependent. Our third finding contributes to this discussion with a controlled comparison across thinking budgets within the same model families.

\textbf{MCQA as an Evaluation Format}
Multiple-choice question answering remains the dominant primitive for LLM leaderboards but carries known artifacts. Robinson and Wingate~\cite{robinson2023leveraging} characterize how models exploit option likelihoods, and Zheng et al.~\cite{zheng2024large} document strong answer-position bias. Pezeshkpour and Hruschka~\cite{pezeshkpour2024large} show that simply permuting option order can change predictions on a substantial fraction of items. We adopt MCQA because it underpins the benchmarks our findings speak to most
directly. The answer-extraction logic is held fixed across all techniques, so that no
technique is advantaged by a parser written for it. We do not resample, i.e., every cell
is a single generation at \texttt{temperature=0}.

\textbf{Dataset Saturation}
The saturation claims in~\cite{chollet2019measure, kiela2021dynabench} have been
challenged by~\cite{mcintosh2024inadequacies} and by harder benchmarks such
as GPQA~\cite{rein2023gpqa} and SuperGPQA~\cite{supergpqa2024}. Results from our work in~\S\ref{sec:difficulty}
adds direct evidence from our 10-dataset suite.
\section{Methodology}
\begin{figure}[t!]
\vspace{-3mm}
    \centering
    \includegraphics[width=0.8\linewidth]{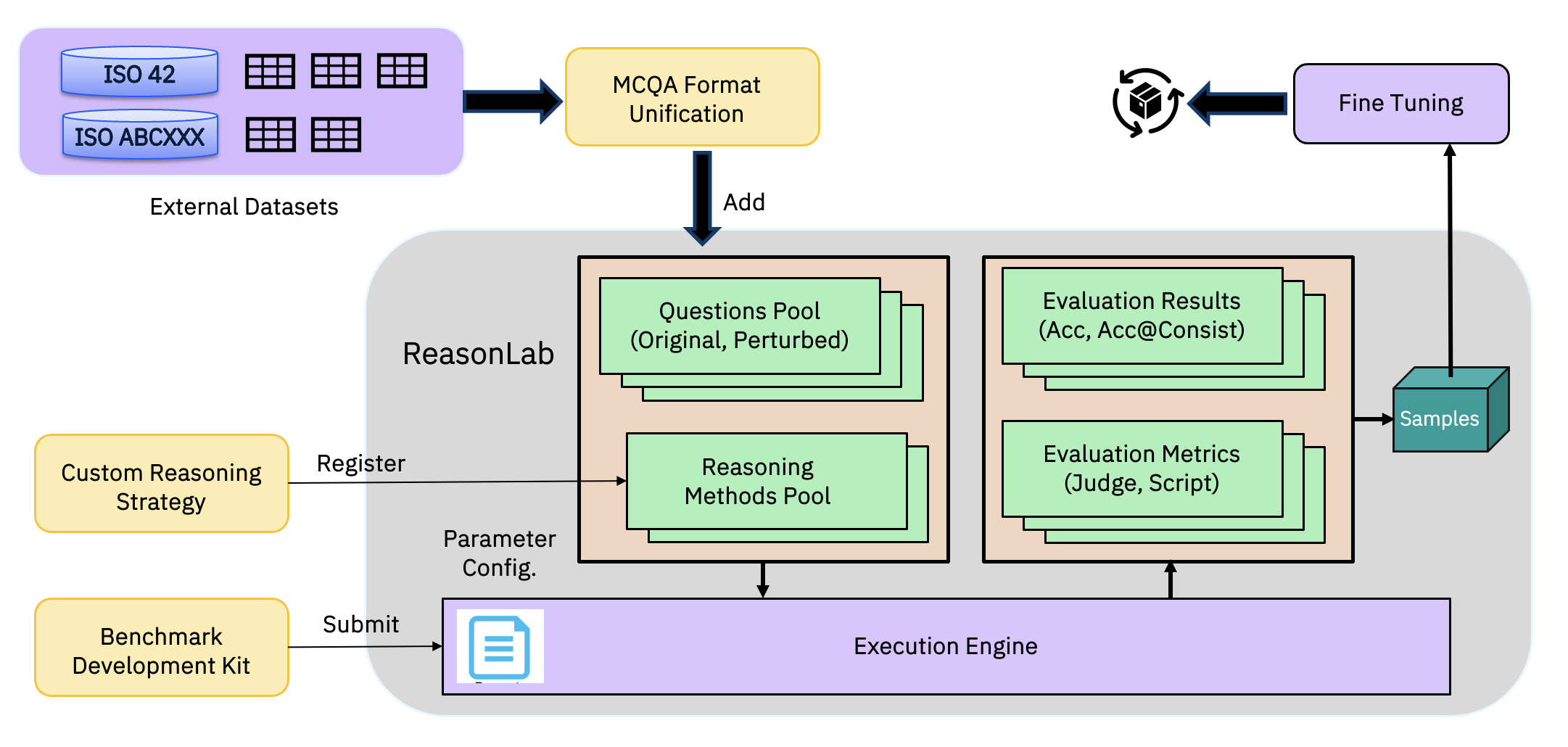}
    \caption{End-to-end ReasonLab evaluation pipeline. A (technique, model, dataset) cell is dispatched to the appropriate prompting module, executed against the model under a single answer-extraction logic, and scored on accuracy and token cost.}
    \label{fig:framework}
\vspace{-3mm}
\end{figure}

\vspace{-3mm}
\label{sec:methodology}
ReasonLab exposes five interchangeable components of the MCQA pipeline, (1) \emph{question presentation}, (2) \emph{context construction}, (3) \emph{reasoning trigger}, (4) \emph{answer generation}, and (5) \emph{answer extraction} so that any prompting technique in the literature can be expressed as a substitution at one of these points (Figure~\ref{fig:framework}). This lets us compare techniques head-to-head under a single protocol rather than inheriting each technique's original protocol.
\subsection{Datasets}
\begin{figure}[h]
\vspace{-1mm}
    \centering
    \includegraphics[width=\linewidth]{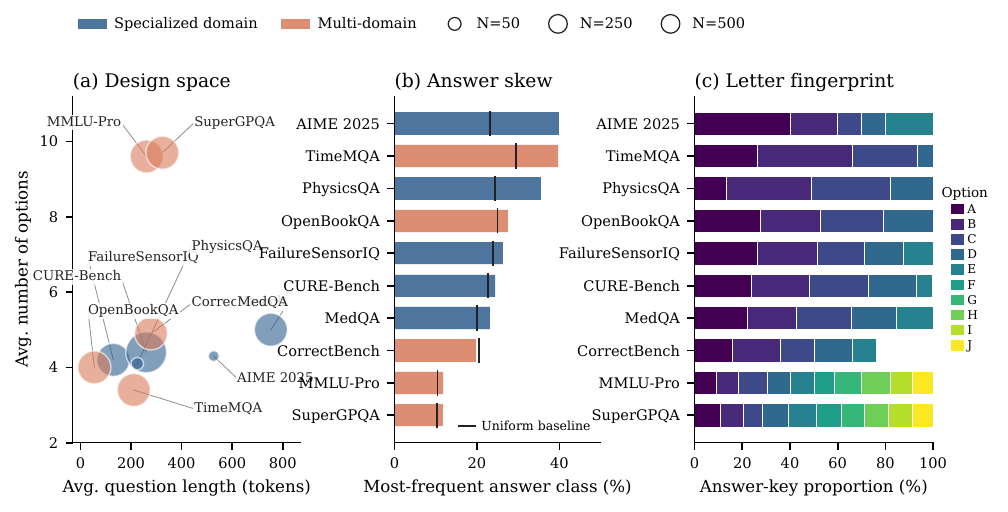}
    \caption{\textbf{Dataset suite at a glance.}
    (a)~\emph{Design space:} mean question length vs.\ mean number of answer options; bubble area is proportional to record count, colour separates single-domain from multi-domain benchmarks.
    (b)~\emph{Answer-distribution skew:} proportion of correct answers in the most-frequent letter class; the black tick marks the uniform baseline ($1/k$ for $k$ options). Bars past the tick indicate answer-key bias.
    (c)~\emph{Letter fingerprint:} full per-letter distribution as a stacked bar; datasets share panel (b)'s ordering, most-skewed first.}
    \label{fig:dataset-suite}
\vspace{-3mm}
\end{figure}

\vspace{-2mm}
\label{sec:datasets}
We evaluate on 10 MCQA datasets spanning physics, medicine, mathematics, IoT/time-series, science, and multi-domain graduate knowledge: \textbf{FailureSensorIQ}~\cite{failuresensoriq}, \textbf{CURE-Bench}~\cite{curebench2024}, \textbf{CorrectBench}~\cite{tie2025correctbench}, \textbf{PhysicsQA}~\cite{physicsqa2023}, \textbf{AIME~2025~QA}~\cite{aime2025qa}, \textbf{MedQA}~\cite{jin2021medqa}, \textbf{MMLU-Pro}~\cite{wang2024mmlupro}, \textbf{OpenBookQA}~\cite{OpenBookQA2018}, \textbf{SuperGPQA}~\cite{supergpqa2024}, and \textbf{Time-MQA}~\cite{kong2025timemqatimeseriesmultitask}. The design space
and the skew of the answer-keys are visualized in Figure~\ref{fig:dataset-suite}.
And per-dataset record counts, average question lengths, option counts, and
full per-letter answer-key distributions are reported in
Appendix Table~\ref{tab:dataset_stats}.

The 10 datasets cover a wide range along three axes that matter for MCQA evaluation, primarily, question length (55--753 tokens on average), number of answer options (2--10), and domain breadth (single-domain such as PhysicsQA to broad multi-domain suites such as MMLU-Pro and SuperGPQA). Six of the ten datasets exhibit answer-key skew above the uniform $1/k$ baseline, which we control for through fixed answer-extraction logic.

\subsection{Models}
\vspace{-2mm}
\label{sec:models}
We evaluated \textbf{27 model configurations} spanning eight families. A configuration is
a model together with the reasoning setting under which it was run. We set that setting in
the harness rather than relying on a vendor default where each configuration declares a
reasoning effort, and the harness converts it into an explicit reasoning-token budget that
is added to the generation limit, namely \emph{nothink} (no reasoning requested),
\emph{low}~=~1{,}024, \emph{medium}~=~8{,}192 and \emph{high}~=~24{,}576 reasoning tokens
on top of a 4{,}096-token content limit. Separating the two limits in this way keeps the
content budget constant while the reasoning budget varies, and thus allows the
contribution of test-time reasoning to be disentangled from the raw model capability.

Two qualifications apply to this control. First, on configurations served without a
reasoning channel the single 4{,}096-token limit governs the whole generation, and so
reasoning and content draw on the same budget. Second, some models like GPT-5
(\texttt{2025-08-07}) and Gemini~2.5~Pro, emit reasoning tokens even when no reasoning
effort is requested. We verify this from the logged token counts rather than from the
requested configuration, and we report the enable-reasoning comparison of
\S\ref{sec:thinking} only for those models whose \emph{nothink} configuration is confirmed
to emit no reasoning tokens.

Two design choices shape this set. First, we prioritize coverage of the major frontier
families (GPT-5, Claude Sonnet~4, Gemini~2.5 Pro, Llama, Qwen3, Mistral, Granite, GPT-OSS)
over depth within any single family. Second, we keep a parameter range from 8B to 405B
disclosed parameters so that the ranking analysis of \S\ref{sec:qwensurprise} is not driven
by a single point.\newline

\textbf{Technique-model application policy.}
All 8 prompting techniques are applied to \textbf{12 of the 27 configurations}, which fall
into two groups that we report separately.

\emph{Group A, no reasoning requested (8 configurations).} Llama-3.2-11B-Vision,
Llama-3.3-70B, Llama-4-Maverick-17B-128e, Mistral-Large-2512, Mistral-Medium-2505,
Mistral-Small-3.1-24B-2503, Granite-3.3-8B and Granite-4-h-small. Here an external
reasoning trigger is the only reasoning the model is asked to perform, and thus the
comparison isolates the effect of the prompt.

\emph{Group B, reasoning enabled (4 configurations).} GPT-OSS-20B, GPT-OSS-120B, Qwen3-8B
and Qwen3-30B-A3B-Thinking-2507. These open-weight models reason natively, and as such an
external trigger is stacked on top of an internal trace rather than substituting for it.
The two groups therefore answer different questions and are not pooled, i.e., Group A
measures whether a prompt can elicit reasoning, whereas Group B measures whether a prompt
adds anything once the model already reasons.

The remaining 15 configurations, all of them vendor-hosted reasoning models, are evaluated
under \textbf{baseline prompting only} (full configuration list in
Appendix~\ref{app:models}, Table~\ref{tab:models}), because we cannot hold their serving
configuration fixed across techniques in the way Group B allows.

\subsection{Prompting techniques}
\vspace{-2mm}
\label{sec:prompting}
ReasonLab implements 8 prompting techniques (Table~\ref{tab:tools}) spanning four families: direct prompting, reasoning triggers, prompt-side robustness, and self-generated context. The two CoT variants (Expert, Inductive) are minor framings of CoT-Standard but yield qualitatively different results (\S\ref{sec:results}).
\begin{table}[ht!]
\vspace{-3mm}
    \centering
    \small
    \caption{The 8 prompting techniques in ReasonLab.}
    \label{tab:tools}
    \setlength{\tabcolsep}{4pt}
    \begin{tabularx}{\linewidth}{@{}lX@{}}
        \toprule
        \textbf{Technique} & \textbf{Trigger \textcolor{blue}{\textcircled{T}} / mechanism} \\
        \midrule
        Baseline & Direct prompt; answer letter only. \\
        CoT-Standard~\cite{wei2022chain} & \textit{``Let me think step by step.''} \textcolor{blue}{\textcircled{T}} \\
        CoT-Expert & \textit{``\ldots as a reliability engineer.''} \textcolor{blue}{\textcircled{T}} \\
        CoT-Inductive & \textit{``\ldots step-by-step inductive reasoning, given the domain-specific nature.''} \textcolor{blue}{\textcircled{T}} \\
        Self-Plan-and-Solve~\cite{wang2023planandsolve} & \textit{``Gather information, devise a plan, answer step by step.''} \textcolor{blue}{\textcircled{T}} \\
        Prompt-Repeat~\cite{promptrepeat} & Direct prompt repeated multiple times. \\
        Self-Generate~\cite{yu2022generate} & Generates contextual document, then conditions on it. \\
        Self-Analogical~\cite{yasunaga2024large} & Generates related MCQA exemplars, then conditions on them. \\
        \bottomrule
    \end{tabularx}
\end{table}
\vspace{-3mm}
\subsection{Raw vs.\ matched comparison}
\vspace{-2mm}
\label{sec:matched}
The application policy creates two natural levels of analysis, and we report both
throughout the paper.

\textbf{Raw comparison.} For each technique, we aggregate accuracy across every (dataset,
model) cell on which it was run. This uses all available data, but is unbalanced, i.e.,
techniques applied to a stronger or weaker subset of models will have inflated or deflated
raw means.\newline
\textbf{Matched comparison.} We restrict each comparison to the (dataset, model) cells
common to all 8 techniques within a group. For Group~A this gives
$8\times10=80$ cells, and for Group~B it gives $4\times10=40$ cells. Within a
group the design is fully balanced, i.e., every technique is evaluated on exactly the same
questions on exactly the same models, and so paired statistical tests across techniques are
licensed. We use the matched comparison as our \textbf{headline result} for this reason.
Reported $p$-values come from paired $t$-tests over the cells of a group, with each cell's
accuracy as the unit of observation, and we report the median difference and a Wilcoxon
signed-rank test alongside the mean, since cell-level differences are not guaranteed to be
symmetric.

\subsection{Evaluation metrics}
\vspace{-2mm}
\label{sec:metrics}
We report \textbf{accuracy} per cell (technique, model, dataset) and \textbf{token cost}
as separate input, output and reasoning-token counts (cost and accuracy frontiers in
\S\ref{sec:thinking}). All experiments use \texttt{temperature=0} and identical system
prompts.\newline
\textbf{Answer extraction} is performed by one procedure shared by every technique, so
that no technique is advantaged by a parser written for it. The prompt requires the answer
as a strict JSON object of the form \texttt{\{"answer": [...]\}}. From the returned text we
take the last JSON object matched by a regular expression, parse it with
\texttt{json.loads}, and fall back to \texttt{json\_repair} when that fails. If no object
can be parsed, the model is re-prompted with a note that the previous attempt did not
produce the requested block, for up to three attempts in total, after which the cell is
scored as incorrect. The last object is taken rather than the first because techniques
that instruct the model to work through exemplars emit intermediate objects in the same
format, and taking the first would score an exemplar in place of the answer.\newline
\textbf{Model comparison.} Where we compare model configurations rather than
techniques (\S\ref{sec:qwensurprise}), we report accuracy with question-level bootstrap
95\% confidence intervals, restricted to configurations that share dataset coverage so
that no configuration is scored on an easier subset of the suite than another.

\section{Results}
\vspace{-3mm}
\label{sec:results}
\subsection{Technique effects under matched comparison}
\vspace{-2mm}
\label{sec:baseline}
\begin{table}[h]
\vspace{-3mm}
\centering
\small
\setlength{\tabcolsep}{4pt}
\caption{\textbf{Matched comparison of the 8 prompting techniques.} Each group is
restricted to the (dataset, model) cells on which all 8 techniques were evaluated, so the
design is balanced within a group. $\Delta$ is the paired mean of (baseline $-$ technique)
and a negative value favours the technique; Med.\ is the paired median. $p_t$ is from a
paired $t$-test and $p_W$ from a Wilcoxon signed-rank test, both over cells, with each
cell's accuracy as the unit of observation. W/T/L counts cells where baseline wins, ties
or loses. Techniques are ordered by mean accuracy within each group.}
\label{tab:techniques}
\begin{tabular}{lrrrrrrr}
\toprule
Technique & Mean & 95\% CI & $\Delta$ (pp) & Med.\ (pp) & $p_t$ & $p_W$ & W/T/L \\
\midrule
\multicolumn{8}{l}{\emph{Group A: no reasoning requested, 8 configurations} (80 cells)} \\
\midrule
SelfAnalogic & 52.96 & $\pm$4.57 & -4.69 & -4.42 & $<\!\!10^{-6}$ & $<\!\!10^{-6}$ & 16/0/64 \\
SelfPlanSolve & 52.88 & $\pm$4.46 & -4.61 & -3.42 & $<\!\!10^{-6}$ & $<\!\!10^{-6}$ & 17/3/60 \\
CoT & 52.68 & $\pm$4.46 & -4.41 & -3.52 & $<\!\!10^{-6}$ & $<\!\!10^{-6}$ & 18/3/59 \\
CoTExpert & 52.52 & $\pm$4.45 & -4.25 & -2.80 & $<\!\!10^{-6}$ & $<\!\!10^{-6}$ & 21/4/55 \\
CoTInductive & 52.19 & $\pm$4.41 & -3.92 & -3.02 & $<\!\!10^{-6}$ & $<\!\!10^{-6}$ & 21/3/56 \\
\textbf{baseline} & \textbf{48.27} & $\pm$4.22 & n/a & n/a & n/a & n/a & n/a \\
SelfGenerateMethod & 47.17 & $\pm$3.96 & +1.09 & +1.80 & 0.075 & 0.006 & 52/2/26 \\
PromptRepeat & 47.12 & $\pm$4.42 & +1.15 & +0.40 & 0.017 & 0.021 & 45/5/30 \\
\midrule
\multicolumn{8}{l}{\emph{Group B: reasoning enabled, 4 configurations} (40 cells)} \\
\midrule
\textbf{baseline} & \textbf{70.84} & $\pm$4.23 & n/a & n/a & n/a & n/a & n/a \\
CoT & 70.69 & $\pm$4.51 & +0.14 & -0.16 & 0.760 & 0.592 & 16/3/21 \\
CoTInductive & 70.67 & $\pm$4.36 & +0.17 & +0.00 & 0.581 & 0.377 & 19/3/18 \\
SelfAnalogic & 70.46 & $\pm$4.33 & +0.37 & +0.60 & 0.466 & 0.219 & 23/1/16 \\
PromptRepeat & 70.45 & $\pm$4.35 & +0.39 & +0.00 & 0.290 & 0.407 & 18/6/16 \\
CoTExpert & 70.33 & $\pm$4.46 & +0.50 & +0.20 & 0.167 & 0.192 & 20/2/18 \\
SelfPlanSolve & 70.32 & $\pm$4.30 & +0.51 & +0.00 & 0.266 & 0.381 & 19/5/16 \\
SelfGenerateMethod & 67.89 & $\pm$4.55 & +2.95 & +2.80 & $7.8\!\times\!10^{-5}$ & $9.9\!\times\!10^{-5}$ & 29/3/8 \\
\bottomrule
\end{tabular}
\end{table}

\begin{figure}[t]
\vspace{-3mm}
    \centering
    \includegraphics[width=\linewidth]{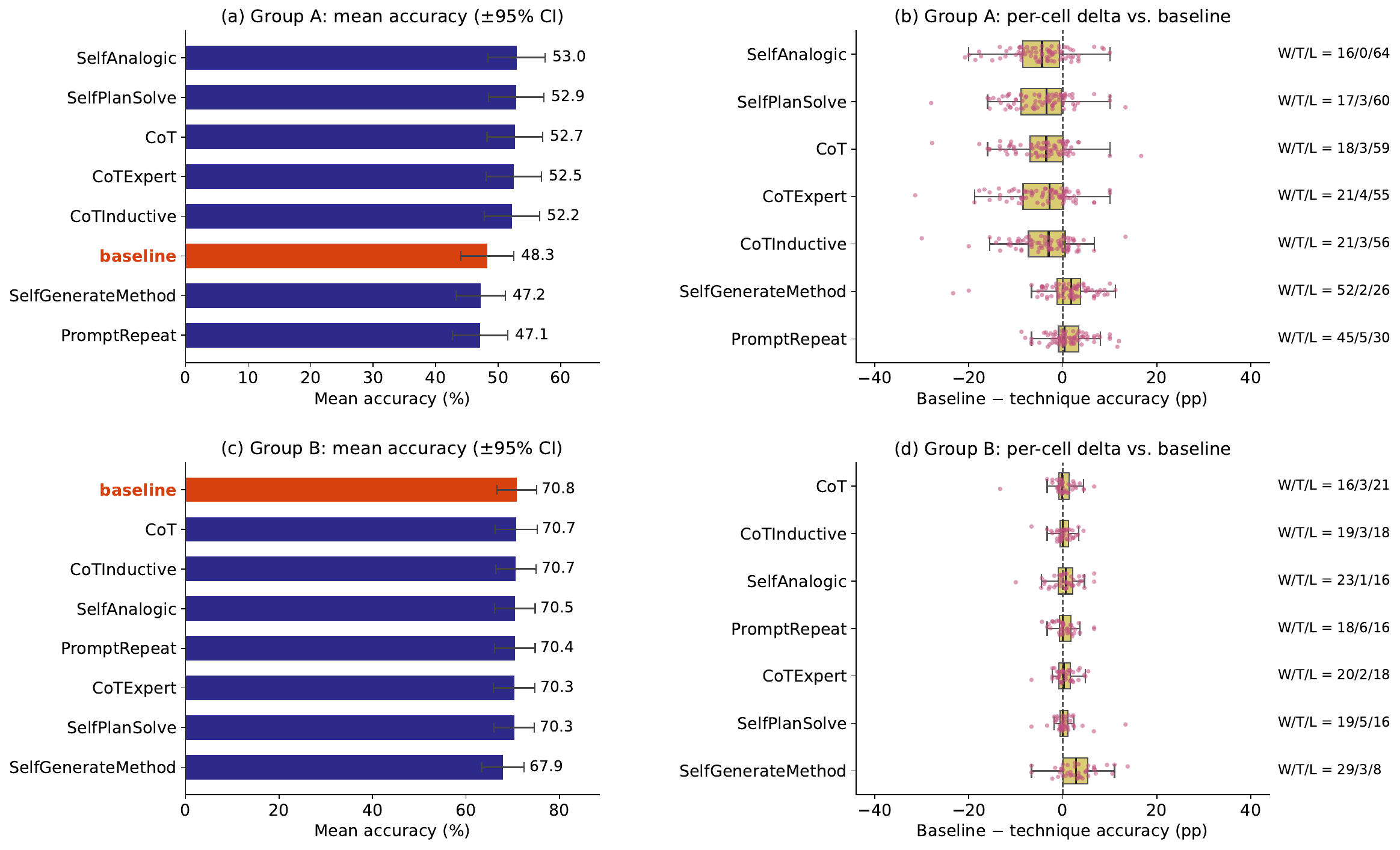}
    \caption{\textbf{Matched comparison of the 8 prompting techniques.}
    (a, b)~Group~A, i.e., the 80 (dataset, model) cells in which every technique was
    evaluated against the same 8 configurations run with no reasoning requested.
    (c, d)~Group~B, i.e., the 40 cells covering the 4 open-weight configurations run with
    reasoning enabled. The left panels give mean accuracy per technique with 95\%
    confidence intervals, with baseline highlighted. The right panels give the per-cell
    distribution of (baseline $-$ technique) accuracy in pp, where positive values
    indicate baseline wins, and W/T/L gives paired wins, ties and losses. Panels (b)
    and (d) share a horizontal scale, so the contraction of the technique effect once the
    model reasons natively is directly visible.}
    \label{fig:matched}
\vspace{-3mm}
\end{figure}

Comparing techniques by their raw means is not informative here, because coverage is
unequal. Baseline was additionally evaluated on the 15 vendor-hosted reasoning
configurations, and its raw mean over all 244 cells on which it was run is 60.79\%,
against 48.27\% on the 80 cells of Group~A. That difference is a property of which models
baseline was run on rather than of the prompt itself. We therefore restrict every
comparison to the cells on which all 8 techniques were evaluated, as described in
\S\ref{sec:matched}. Table~\ref{tab:techniques} reports the result for both groups and
Figure~\ref{fig:matched} shows the corresponding means and per-cell differences.

\textbf{Group~A.} On the 8 configurations run with no reasoning requested, five
reasoning triggers improve on direct prompting, namely Self-Analogical (52.96\%,
$\Delta=-4.69$~pp), Plan-and-Solve (52.88\%, $-4.61$~pp), CoT (52.68\%, $-4.41$~pp),
CoT-Expert (52.52\%, $-4.25$~pp) and CoT-Inductive (52.19\%, $-3.92$~pp). All five are
significant at $p<10^{-6}$ under both the paired $t$-test and the Wilcoxon signed-rank
test, each wins between 55 and 64 of the 80 cells, and for each the mean and the median
difference agree in sign and magnitude. However, the five are separated by only 0.77~pp in
total, i.e., which trigger is used matters far less than whether one is present at all.
The two techniques that are not reasoning triggers behave differently, i.e., Prompt-Repeat
and Self-Generate sit approximately 1.1~pp \emph{below} baseline, and neither reaches
significance under the paired $t$-test. The separation in Table~\ref{tab:techniques} is
therefore between prompts that ask the model to reason and prompts that do not, rather
than between simple and elaborate scaffolds.

\textbf{Group~B.}
\label{sec:armb}
On the 4 open-weight configurations that reason natively, the effect disappears. No
technique differs from direct prompting by more than 0.51~pp, and none of the differences
approaches significance under either test ($p\geq0.17$ paired $t$, $p\geq0.19$ Wilcoxon).
The single exception is Self-Generate, at 2.95~pp below baseline, significant at
$p<10^{-4}$ under both tests and losing 29 of 40 cells. Read together with Group~A, this gives a
consistent account, i.e., a reasoning trigger is worth roughly four points when the model
is not already reasoning, and nothing measurable once it is. Self-Generate is the only
technique that is reliably harmful in either group, which is consistent with it being the
only technique that replaces the question context rather than adding an instruction to it.

\begin{figure}[t]
\vspace{-3mm}
    \centering
    \includegraphics[width=\linewidth]{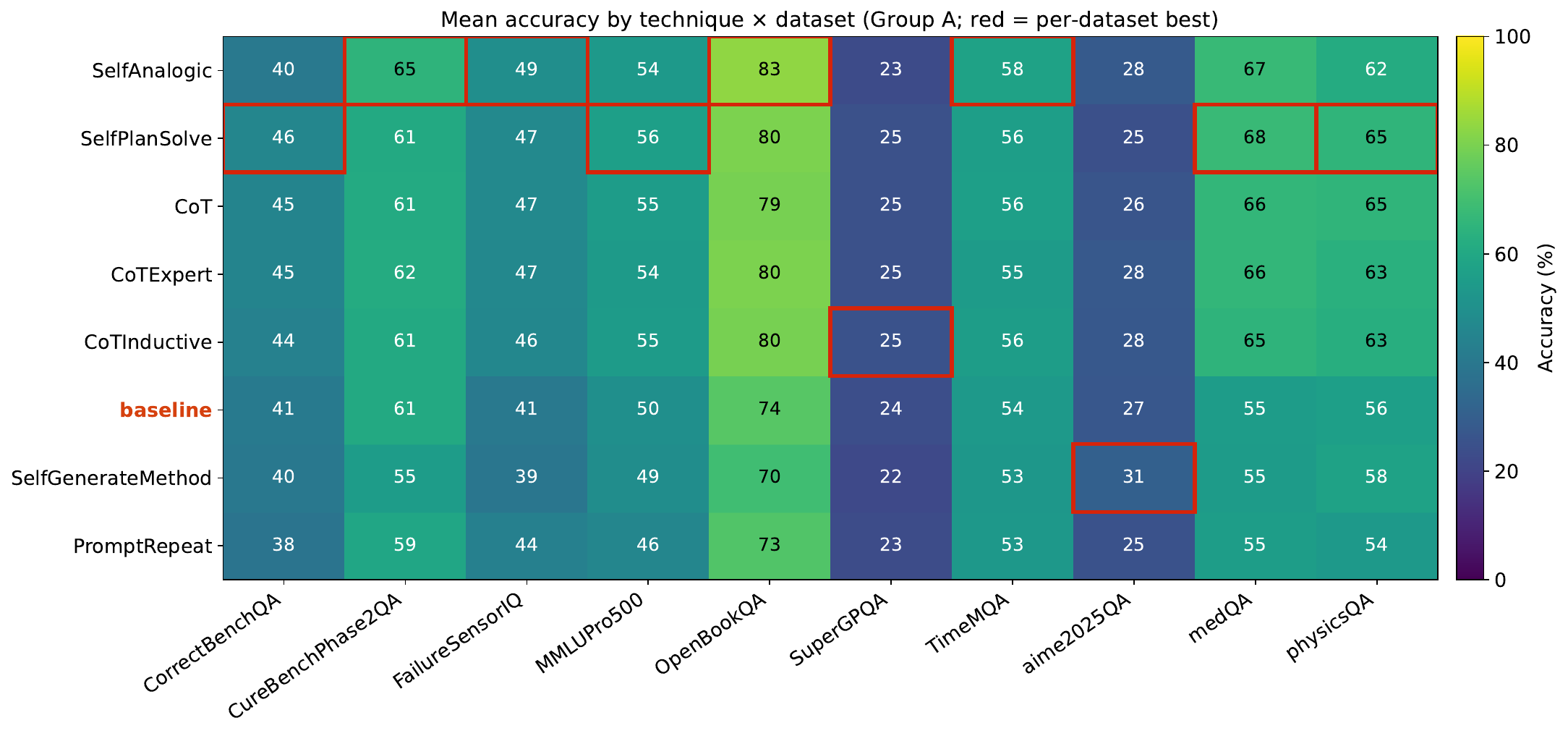}
    \caption{\textbf{Mean accuracy by technique $\times$ dataset on the Group~A matched
    set.} The per-dataset best technique is outlined in red. Self-Analogical wins 4
    datasets, Plan-and-Solve 3, and CoT, CoT-Inductive and Self-Generate one each, while
    baseline wins none outright.}
    \label{fig:heatmap}
\vspace{-3mm}
\end{figure}

The per-dataset structure, summarised in Figure~\ref{fig:heatmap}, supports the same
reading. On Group~A no single technique dominates, i.e., Self-Analogical wins 4 of the
10 datasets, Plan-and-Solve wins 3, and CoT, CoT-Inductive and Self-Generate win one each,
while baseline wins none outright. On Group~B, where no technique separates from baseline overall, the per-dataset
winners are correspondingly scattered, i.e., Self-Analogical takes 3 of the 10 datasets,
baseline and Prompt-Repeat 2 each, and CoT, CoT-Expert and Self-Generate one each, with
every margin below 1.5~pp. The one
dataset on which a self-generated-context technique wins in both groups is AIME~2025,
where Self-Generate leads, although with only 30 questions that result should be treated
as indicative rather than established.

\subsection{Model comparison under matched coverage}
\begin{figure}[t]
\vspace{-2mm}
    \centering
    \includegraphics[width=.85\linewidth]{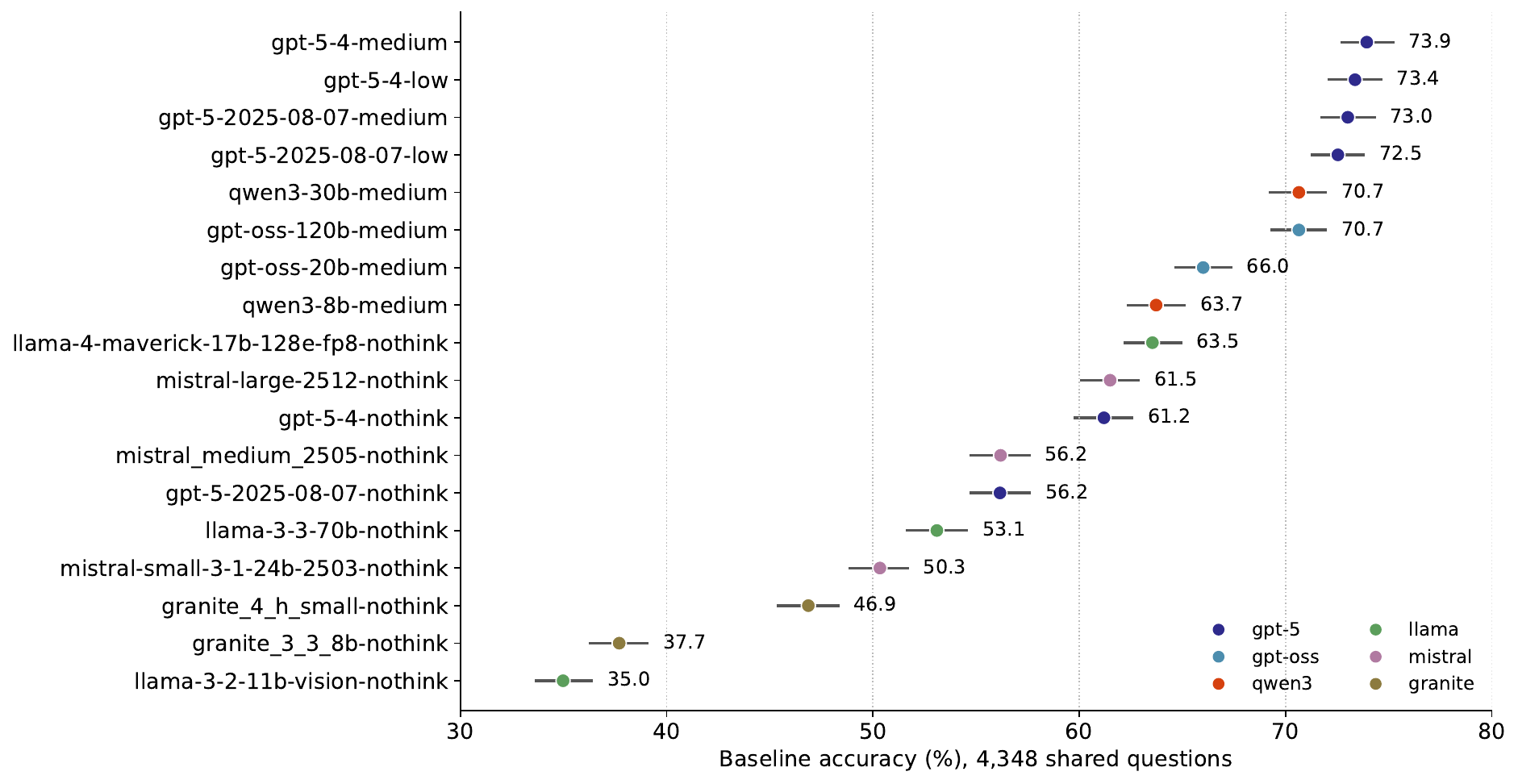}
    \caption{\textbf{Baseline accuracy per configuration with question-level bootstrap
    95\% confidence intervals}, restricted to the 18 configurations evaluated on
    all 10 datasets. Colour denotes model family.}
    \label{fig:accuracy}
\vspace{-3mm}
\end{figure}

\vspace{-2mm}
\label{sec:qwensurprise}
Configurations in our suite were not all evaluated on the same datasets, and mean accuracy
over different dataset mixes is not comparable. We therefore compare models only on the
18 configurations evaluated on all 10 datasets, which share
4{,}348 questions, and we attach a 95\% confidence interval obtained by
resampling those questions with replacement (2{,}000 replicates, a common resample across
configurations). Configurations with partial coverage, notably the Claude~Sonnet~4.6
variants, are excluded from this comparison rather than ranked against a fuller suite;
listing them alongside would reward a configuration for having been evaluated on the
easier half of the benchmark. Table~\ref{tab:accuracy_ci} and Figure~\ref{fig:accuracy}
report the result.
\begin{table}[h]
\vspace{-2mm}
\centering
\small
\caption{\textbf{Baseline accuracy with question-level bootstrap 95\% confidence
intervals}, for the 18 configurations evaluated on all 10 datasets. All
configurations answer the same 4{,}348 questions, so the intervals are directly
comparable.}
\label{tab:accuracy_ci}
\begin{tabular}{lrc}
\toprule
Configuration & Accuracy (\%) & 95\% CI \\
\midrule
gpt-5-4-medium & 73.94 & [72.68, 75.28] \\
gpt-5-4-low & 73.37 & [72.03, 74.70] \\
gpt-5-2025-08-07-medium & 73.02 & [71.69, 74.38] \\
gpt-5-2025-08-07-low & 72.54 & [71.23, 73.85] \\
gpt-oss-120b-medium & 70.65 & [69.27, 72.01] \\
qwen3-30b-medium & 70.65 & [69.20, 72.01] \\
gpt-oss-20b-medium & 66.01 & [64.63, 67.43] \\
qwen3-8b-medium & 63.73 & [62.30, 65.18] \\
llama-4-maverick-17b-128e-fp8-nothink & 63.55 & [62.17, 65.00] \\
mistral-large-2512-nothink & 61.50 & [60.05, 62.95] \\
gpt-5-4-nothink & 61.20 & [59.73, 62.63] \\
mistral\_medium\_2505-nothink & 56.19 & [54.69, 57.64] \\
gpt-5-2025-08-07-nothink & 56.16 & [54.69, 57.66] \\
llama-3-3-70b-nothink & 53.10 & [51.61, 54.62] \\
mistral-small-3-1-24b-2503-nothink & 50.34 & [48.83, 51.75] \\
granite\_4\_h\_small-nothink & 46.87 & [45.35, 48.37] \\
granite\_3\_3\_8b-nothink & 37.70 & [36.22, 39.14] \\
llama-3-2-11b-vision-nothink & 34.98 & [33.62, 36.43] \\
\bottomrule
\end{tabular}
\end{table}

Two observations follow. First, the ordering is not a ranking in any strong sense, i.e.,
the four GPT-5 configurations span 72.54\% to 73.94\% with overlapping intervals, and
GPT-OSS-120B and Qwen3-30B-A3B-Thinking-2507 are indistinguishable at 70.65\% each. Only
differences of roughly 2~pp or more are resolved at this sample size, and we therefore
describe groups of configurations rather than assigning ranks.

Second, disclosed parameter count does not order the configurations. Qwen3-30B-A3B
activates approximately 3B parameters per token and matches GPT-OSS-120B, which activates
approximately 5.1B of 120B total, while clearly exceeding GPT-OSS-20B (66.01\%). The
comparison is sharpest within the Llama family, where all three configurations run without
reasoning and share 9 datasets and 3{,}569 questions: Llama-4-Maverick, with 17B active
parameters, reaches 62.40\% [60.75, 64.00] against 52.37\% [50.57, 54.05] for
Llama-3.1-405B and 50.41\% [48.78, 52.12] for Llama-3.3-70B. The intervals do not overlap,
and on the questions the two answer in common Maverick is correct where Llama-3.1-405B is
wrong on 645 questions against 287 the other way. Architecture, training data and recency
dominate disclosed scale within this suite. We make this claim comparatively and do not
extrapolate it into a scaling law, since our suite fixes the task format and the decoding
protocol.

\subsection{Reasoning budgets: a steep step and a flat plateau}
\begin{figure}[h]
\vspace{-2mm}
    \centering
    \includegraphics[width=\linewidth]{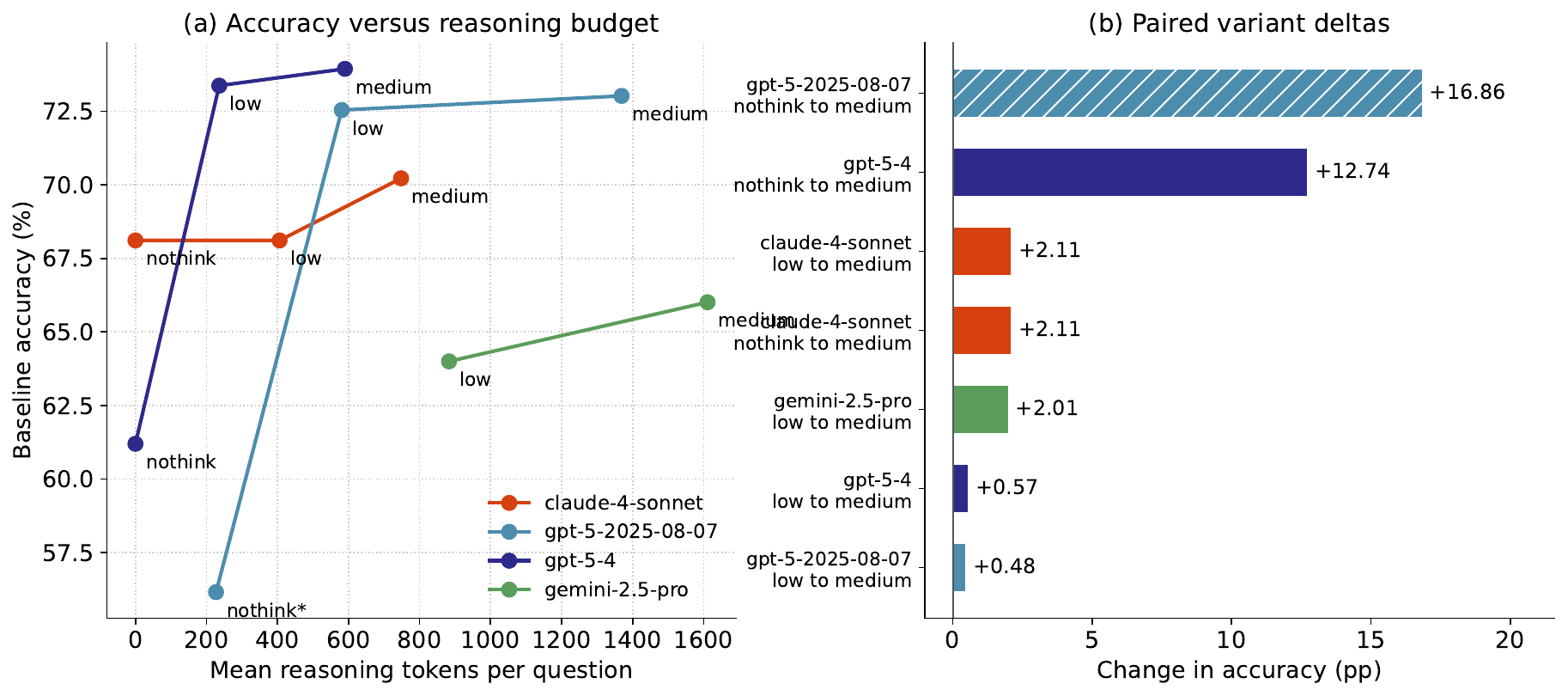}
    \caption{\textbf{Reasoning-budget analysis, within model family, on the datasets where
    every variant of that family was evaluated.}
    (a)~Accuracy against mean reasoning tokens per question. A \emph{nothink} point marked
    with an asterisk still emitted reasoning tokens on some questions.
    (b)~Paired variant deltas; the hatched bar is the one enable-reasoning contrast whose
    \emph{nothink} baseline is not clean.}
    \label{fig:thinking}
\vspace{-3mm}
\end{figure}

\vspace{-2mm}
\label{sec:thinking}
We analyse the reasoning control on the four model families that have multiple variants on
a comparable dataset slice, namely Claude Sonnet~4, Gemini~2.5~Pro, GPT-5
(\texttt{2025-08-07}) and GPT-5-4. For each family we restrict to the datasets on which
every variant of that family was evaluated, compare within model, and pair over the shared
questions. Figure~\ref{fig:thinking} reports the resulting accuracy and cost curves
together with the paired variant deltas.

Before reading the deltas, one property of the \emph{nothink} setting has to be
established rather than assumed, because a request for no reasoning is not always honoured.
We check it in the logs. Claude Sonnet~4 and GPT-5-4 emit \textbf{zero} reasoning tokens
under \emph{nothink} on every question, so for those two families the
\emph{nothink}~$\rightarrow$~\emph{medium} contrast is a genuine off-to-on comparison.
GPT-5 (\texttt{2025-08-07}) emits reasoning tokens on 18.2\% of questions under the same
setting, averaging 1{,}369 tokens on the questions where it does reason, and Gemini~2.5~Pro
offers no setting that suppresses reasoning at all. We report those two families for
budget scaling only, and we exclude them from the enable-reasoning claim.

\textbf{Enabling reasoning is the dominant lever, and its value is model-dependent.} On the
two families with a verified \emph{nothink}, switching to \emph{medium} adds
\textbf{+12.74~pp} for GPT-5-4 (73.94\% against 61.20\%, paired over 4{,}348 questions) and
\textbf{+2.10~pp} for Claude Sonnet~4 (70.22\% against 68.11\%, 3{,}569 questions). The
same switch on GPT-5 (\texttt{2025-08-07}) measures $+16.86$~pp, but as noted that
comparison is partial-reasoning against full reasoning rather than off against on, and we
do not count it as an enable-reasoning effect. The spread between GPT-5-4 and Claude is
itself the finding, i.e., the same intervention is worth six times as much on one model as
on the other, and thus ``enable reasoning'' is not a single quantity that transfers across
models.

\textbf{Scaling the budget within the on-state buys very little.} An eightfold increase in
the reasoning budget, from \emph{low} (1{,}024 tokens) to \emph{medium} (8{,}192 tokens),
adds \textbf{+0.48~pp} for GPT-5 (\texttt{2025-08-07}), \textbf{+0.57~pp} for GPT-5-4,
\textbf{+2.02~pp} for Gemini~2.5~Pro and \textbf{+2.10~pp} for Claude Sonnet~4, while mean
reasoning consumption rises by 342 to 788 tokens per question. The most direct illustration
is Claude Sonnet~4 at \emph{low}, which spends 406 reasoning tokens per question and scores
68.11\%, i.e., exactly what it scores while spending none. On that model the small budget
is paid for and returns nothing measurable.

One reason the plateau is so flat is that the budget is not the binding constraint. Mean
reasoning consumption at \emph{medium} ranges from 590 tokens (GPT-5-4) to 1{,}611
(Gemini~2.5~Pro), i.e., between 7\% and 20\% of the 8{,}192-token budget those
configurations were given. Raising a cap that models are not reaching cannot be expected to
help, and thus the framing of test-time compute as a single dial conflates two distinct
interventions: enabling reasoning, which is large and model-dependent, and enlarging a
budget that is mostly unspent, which is neither.

\subsection{Dataset difficulty: substantial headroom remains}
\begin{figure}[h]
\vspace{-2mm}
    \centering
    \includegraphics[width=\linewidth]{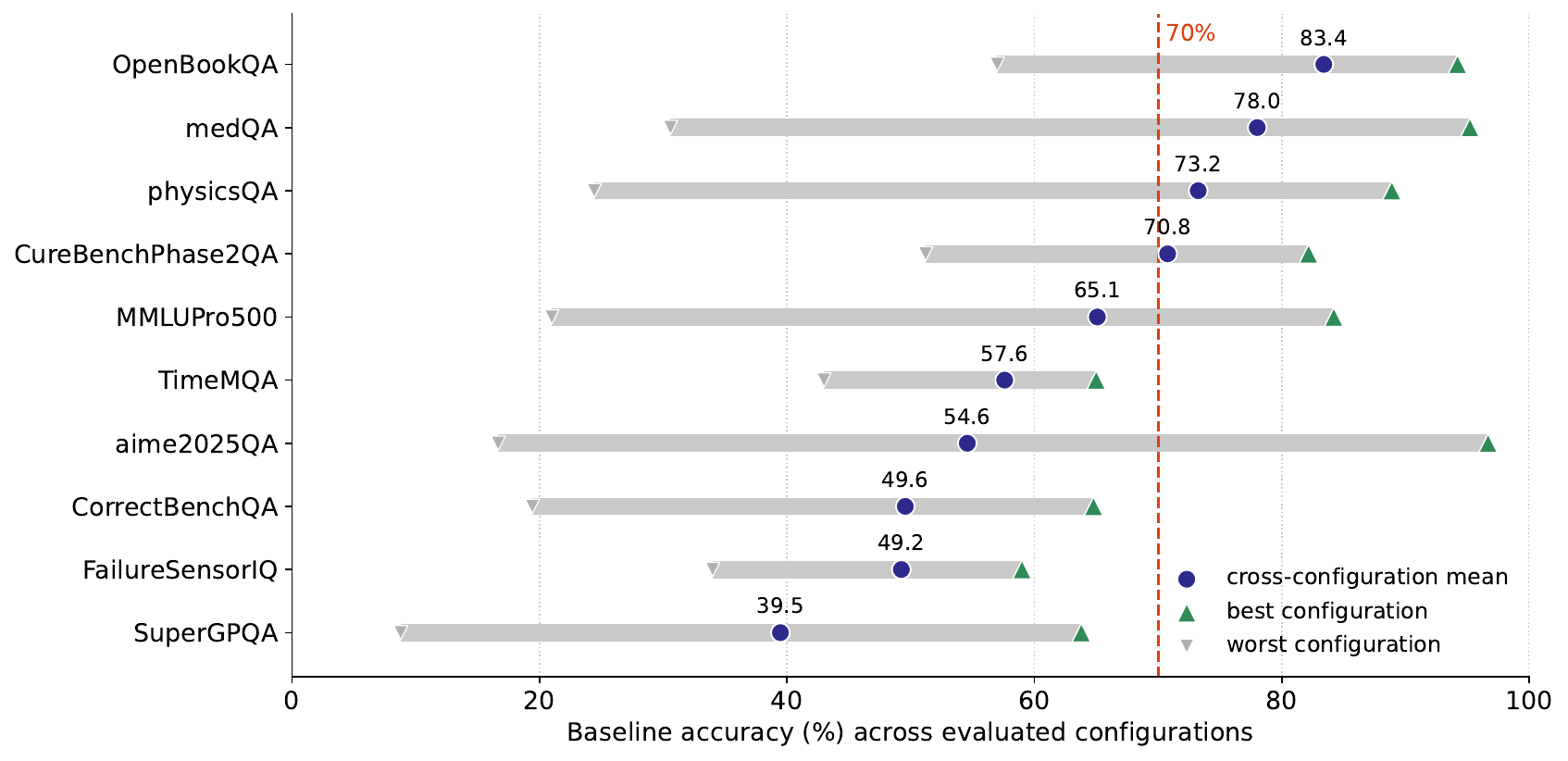}
    \caption{\textbf{Dataset difficulty under baseline prompting.} Each row shows the range
    of accuracies across the configurations evaluated on that dataset (grey bar), the
    cross-configuration mean (circle), and the best and worst configurations (triangles).
    The dashed line marks 70\%. Six datasets have a mean below 70\% and four remain below
    it even for their best configuration.}
    \label{fig:difficulty}
\vspace{-3mm}
\end{figure}

\vspace{-2mm}
\label{sec:difficulty}
Figure~\ref{fig:difficulty} reports the distribution of baseline-prompted accuracy on each
of our 10 datasets, taken across all configurations evaluated on that dataset. Two summary
statistics are worth highlighting. First, the spread from the easiest to the hardest
dataset is large, i.e., OpenBookQA reaches a cross-configuration mean of 83.39\% while
SuperGPQA reaches only 39.48\%, a gap of 43.9~pp. Second, headroom remains substantial,
i.e., 6 of the 10 datasets have a mean below 70\%, and 4 of 10 stay below 70\% even for the
best configuration on each dataset, namely Time-MQA (65.00\%), SuperGPQA (63.80\%),
CorrectBenchQA (64.78\%) and FailureSensorIQ (59.00\%).
This pattern is informative. The hardest benchmarks in our suite, namely SuperGPQA
(graduate-level multi-domain), FailureSensorIQ (industrial sensor reasoning) and
CorrectBenchQA, are precisely the ones where no prompting technique rescues performance,
and \S\ref{sec:baseline} showed that the best available technique moves accuracy on them by
a few points at most. SuperGPQA tops out at 63.80\% even for the strongest configuration we
evaluated.

\textbf{Contamination strengthens rather than weakens this claim.} Several of our datasets
predate the training cutoffs of the models we evaluate, and OpenBookQA, the oldest and the
most likely to appear in pre-training corpora, is also the easiest at 83.39\%. We do not
treat any dataset as contamination-free, and we note that AIME~2025 was released in
February 2025, which is before the cutoff of several configurations in our suite, so we
make no claim that it postdates them. The direction of the bias is what matters here.
Contamination inflates measured accuracy, and therefore the accuracies reported above are
upper bounds on genuine capability. A headroom claim derived from inflated scores is
conservative, i.e., if contamination were removed the measured accuracies would fall and
the headroom would grow. Reports of MCQA saturation therefore appear premature within the
difficulty regime our benchmarks span, and the remaining headroom sits in genuine
reasoning, specialised knowledge and temporal understanding rather than in prompt
engineering.

We defer extended discussion of how to rethink test-time ``thinking'' to
Appendix~\ref{app:discussion}.
\section{Limitations}
\vspace{-3mm}
\label{sec:limitations}
\textbf{Answer extraction is a design choice, not a neutral measurement.} A single
procedure is applied to every technique so that none is advantaged by a parser written for
it, but the choice is consequential. Techniques that instruct the model to work through
exemplars emit several JSON objects in the requested format, and reading the first rather
than the last changes Self-Analogical's measured accuracy on our matched set by more than
30~pp while leaving every other technique within 0.2~pp. Any study that compares prompting
techniques under one extractor inherits this sensitivity, and we release the per-cell
records so the choice can be audited rather than trusted.

\textbf{Unequal dataset coverage restricts the model comparison.}
\S\ref{sec:qwensurprise} is limited to the 18 configurations evaluated on all 10 datasets.
The Claude Sonnet~4.6 variants, which reach the highest accuracy we observe on the subset
they were run on, are therefore absent from that comparison. We prefer a smaller
comparable set to a larger incomparable one, but the omission is real and the suite would
be stronger with uniform coverage.

\textbf{Technique implementation choices.} Each technique is implemented following its
original paper's public template; we did not tune exemplar count, prompt wording, or output
format per model. Self-Analogical in particular was originally validated on mathematical
reasoning under different exemplar settings, and its template carries wording specific to
that original domain into all 10 of ours (Appendix~\ref{app:selfanalog}). Our numbers are
therefore lower bounds on what a tuned implementation of each technique could achieve.

\textbf{Scope and protocol.} Our findings concern multiple-choice question answering and do
not extend directly to open-ended generation or agentic tasks. We use \texttt{temperature=0}
with a single generation per cell, which measures no within-cell variance and excludes
techniques that require multiple samples or search over partial solutions.

\section{Conclusion}
\vspace{-3mm}
Across 480{,}927 model-question evaluations spanning 8 prompting techniques, 10 datasets
and 27 model configurations, three scoped claims survive controlled comparison. First, the
choice of prompting technique is a minor determinant of MCQA accuracy: on configurations
run without a reasoning budget every reasoning trigger we tested improves on direct
prompting by 3.92 to 4.69~pp and none of the triggers separates from the others, while on
configurations that reason natively no technique moves accuracy by more than 0.51~pp. The
one technique with a consistent effect, Self-Generate, reduces accuracy in both regimes.
Second, within our suite and under matched dataset coverage, disclosed parameter count does
not order model configurations, most clearly inside the Llama family where a 17B-active
model exceeds a 405B one by 10~pp with non-overlapping confidence intervals. Third, the
dominant lever in test-time compute is enabling reasoning at all rather than scaling its
budget, and the size of that step is itself model-dependent, ranging from +2.10~pp to
+12.74~pp on the two families where the no-reasoning setting is verified.

Taken together these results argue for redirecting effort from prompt scaffolding toward
base-model capability and harder evaluations. The 43.9~pp spread between the easiest and
hardest benchmark in our suite, and the four datasets that remain below 70\% even for their
best configuration, indicate that the headroom is in the models and the benchmarks rather
than in the prompt.

\bibliographystyle{plainnat}
\bibliography{sample-base}

\begin{thebibliography}{46}
\providecommand{\natexlab}[1]{#1}
\providecommand{\url}[1]{\texttt{#1}}
\expandafter\ifx\csname urlstyle\endcsname\relax
  \providecommand{\doi}[1]{doi: #1}\else
  \providecommand{\doi}{doi: \begingroup \urlstyle{rm}\Url}\fi

\bibitem[{Anthropic}(2025)]{claude4card}
{Anthropic}.
\newblock System card: {Claude Opus 4} \& {Claude Sonnet 4}.
\newblock \url{https://www.anthropic.com/claude-4-system-card}, May 2025.

\bibitem[Balunović et~al.(2025)Balunović, Dekoninck, Petrov, Jovanović, and
  Vechev]{aime2025qa}
Mislav Balunović, Jasper Dekoninck, Ivo Petrov, Nikola Jovanović, and Martin
  Vechev.
\newblock Matharena: Evaluating llms on uncontaminated math competitions,
  February 2025.
\newblock URL \url{https://matharena.ai/}.

\bibitem[Chollet(2019)]{chollet2019measure}
Fran{\c{c}}ois Chollet.
\newblock On the measure of intelligence.
\newblock \emph{arXiv preprint arXiv:1911.01547}, 2019.

\bibitem[Constantinides et~al.(2025)Constantinides, Patel, Lin, Guerrero,
  Patil, and Kalagnanam]{failuresensoriq}
Christodoulos Constantinides, Dhaval Patel, Shuxin Lin, Claudio Guerrero,
  Sunil~Dagajirao Patil, and Jayant Kalagnanam.
\newblock Failuresensoriq: A multi-choice qa dataset for understanding sensor
  relationships and failure modes, 2025.
\newblock URL \url{https://arxiv.org/abs/2506.03278}.

\bibitem[{DeepSeek-AI} et~al.(2025){DeepSeek-AI}, Guo,
  et~al.]{deepseekai2025r1}
{DeepSeek-AI}, Daya Guo, et~al.
\newblock {DeepSeek-R1}: Incentivizing reasoning capability in {LLMs} via
  reinforcement learning.
\newblock \emph{arXiv preprint arXiv:2501.12948}, 2025.

\bibitem[Gao et~al.(2024)Gao, Tow, Abbasi, Biderman, Black, DiPofi, Foster,
  Golding, Hsu, Le~Noac'h, Li, McDonell, Muennighoff, Ociepa, Phang, Reynolds,
  Schoelkopf, Skowron, Sutawika, Tang, Thite, Wang, Wang, and
  Zou]{eval-harness}
Leo Gao, Jonathan Tow, Baber Abbasi, Stella Biderman, Sid Black, Anthony
  DiPofi, Charles Foster, Laurence Golding, Jeffrey Hsu, Alain Le~Noac'h,
  Haonan Li, Kyle McDonell, Niklas Muennighoff, Chris Ociepa, Jason Phang,
  Laria Reynolds, Hailey Schoelkopf, Aviya Skowron, Lintang Sutawika, Eric
  Tang, Anish Thite, Ben Wang, Kevin Wang, and Andy Zou.
\newblock The language model evaluation harness, July 2024.
\newblock URL \url{https://zenodo.org/records/12608602}.

\bibitem[{Gemini Team, Google}(2025)]{gemini25}
{Gemini Team, Google}.
\newblock {Gemini 2.5}: Pushing the frontier with advanced reasoning,
  multimodality, long context, and next generation agentic capabilities, 2025.
\newblock URL \url{https://arxiv.org/abs/2507.06261}.

\bibitem[Hendrycks et~al.(2020)Hendrycks, Burns, Basart, Zou, Mazeika, Song,
  and Steinhardt]{hendrycks2020measuring}
Dan Hendrycks, Collin Burns, Steven Basart, Andy Zou, Mantas Mazeika, Dawn
  Song, and Jacob Steinhardt.
\newblock Measuring massive multitask language understanding.
\newblock \emph{arXiv preprint arXiv:2009.03300}, 2020.

\bibitem[Jin et~al.(2021)Jin, Pan, Oufattole, Weng, Fang, and
  Szolovits]{jin2021medqa}
Di~Jin, Eileen Pan, Nassim Oufattole, Wei-Hung Weng, Hanyi Fang, and Peter
  Szolovits.
\newblock What disease does this patient have? a large-scale open domain
  question answering dataset from medical exams.
\newblock \emph{Applied Sciences}, 11\penalty0 (14):\penalty0 6421, 2021.

\bibitem[Kiela et~al.(2021)Kiela, Bartolo, Nie, Kaushik, Geiger, Wu, Vidgen,
  Prasad, Singh, Ringshia, Ma, Thrush, Riedel, Waseem, Stenetorp, Jia, Bansal,
  Potts, and Williams]{kiela2021dynabench}
Douwe Kiela, Max Bartolo, Yixin Nie, Divyansh Kaushik, Atticus Geiger,
  Zhengxuan Wu, Bertie Vidgen, Grusha Prasad, Amanpreet Singh, Pratik Ringshia,
  Zhiyi Ma, Tristan Thrush, Sebastian Riedel, Zeerak Waseem, Pontus Stenetorp,
  Robin Jia, Mohit Bansal, Christopher Potts, and Adina Williams.
\newblock {Dynabench}: Rethinking benchmarking in {NLP}.
\newblock In \emph{Proceedings of the 2021 Conference of the North American
  Chapter of the Association for Computational Linguistics: Human Language
  Technologies (NAACL-HLT)}, pages 4110--4124. Association for Computational
  Linguistics, 2021.
\newblock \doi{10.18653/v1/2021.naacl-main.324}.

\bibitem[Kojima et~al.(2022)Kojima, Gu, Reid, Matsuo, and
  Iwasawa]{kojima2022large}
Takeshi Kojima, Shixiang~Shane Gu, Machel Reid, Yutaka Matsuo, and Yusuke
  Iwasawa.
\newblock Large language models are zero-shot reasoners.
\newblock \emph{Advances in Neural Information Processing Systems},
  35:\penalty0 22199--22213, 2022.

\bibitem[Kong et~al.(2025)Kong, Yang, Hwang, Du, Zohren, Wang, Jin, and
  Wen]{kong2025timemqatimeseriesmultitask}
Yaxuan Kong, Yiyuan Yang, Yoontae Hwang, Wenjie Du, Stefan Zohren, Zhangyang
  Wang, Ming Jin, and Qingsong Wen.
\newblock Time-mqa: Time series multi-task question answering with context
  enhancement, 2025.
\newblock URL \url{https://arxiv.org/abs/2503.01875}.

\bibitem[Leviathan et~al.(2025)Leviathan, Kalman, and Matias]{promptrepeat}
Yaniv Leviathan, Matan Kalman, and Yossi Matias.
\newblock Prompt repetition improves non-reasoning llms.
\newblock \emph{arXiv preprint arXiv:2512.14982}, 2025.
\newblock URL \url{https://arxiv.org/abs/2512.14982}.

\bibitem[Liang et~al.(2022)Liang, Bommasani, Lee, Tsipras, Soylu, Yasunaga,
  Zhang, Narayanan, Wu, Kumar, et~al.]{liang2022holistic}
Percy Liang, Rishi Bommasani, Tony Lee, Dimitris Tsipras, Dilara Soylu,
  Michihiro Yasunaga, Yian Zhang, Deepak Narayanan, Yuhuai Wu, Ananya Kumar,
  et~al.
\newblock Holistic evaluation of language models.
\newblock \emph{arXiv preprint arXiv:2211.09110}, 2022.

\bibitem[Lu et~al.(2022)Lu, Bartolo, Moore, Riedel, and
  Stenetorp]{lu2022fantastically}
Yao Lu, Max Bartolo, Alastair Moore, Sebastian Riedel, and Pontus Stenetorp.
\newblock Fantastically ordered prompts and where to find them: Overcoming
  few-shot prompt order sensitivity.
\newblock In \emph{Proceedings of the 60th Annual Meeting of the Association
  for Computational Linguistics (Volume 1: Long Papers)}, pages 8086--8098,
  Dublin, Ireland, 2022. Association for Computational Linguistics.
\newblock \doi{10.18653/v1/2022.acl-long.556}.

\bibitem[{M-A-P Team} et~al.(2025){M-A-P Team}, Du, Yao, Ma, Wang, Zheng, Zhu,
  et~al.]{supergpqa2024}
{M-A-P Team}, Xinrun Du, Yifan Yao, Kaijing Ma, Bingli Wang, Tianyu Zheng, Kang
  Zhu, et~al.
\newblock {SuperGPQA}: Scaling llm evaluation across 285 graduate disciplines.
\newblock In \emph{Advances in Neural Information Processing Systems Datasets
  and Benchmarks Track}, 2025.
\newblock arXiv:2502.14739.

\bibitem[McIntosh et~al.(2024)McIntosh, Susnjak, Liu, Watters, and
  Halgamuge]{mcintosh2024inadequacies}
Timothy~R. McIntosh, Teo Susnjak, Tong Liu, Paul Watters, and Malka~N.
  Halgamuge.
\newblock Inadequacies of large language model benchmarks in the era of
  generative artificial intelligence.
\newblock \emph{arXiv preprint arXiv:2402.09880}, 2024.

\bibitem[Mihaylov et~al.(2018)Mihaylov, Clark, Khot, and
  Sabharwal]{OpenBookQA2018}
Todor Mihaylov, Peter Clark, Tushar Khot, and Ashish Sabharwal.
\newblock Can a suit of armor conduct electricity? a new dataset for open book
  question answering.
\newblock In \emph{EMNLP}, 2018.

\bibitem[Mizrahi et~al.(2024)Mizrahi, Kaplan, Malkin, Dror, Shahaf, and
  Stanovsky]{mizrahi2024state}
Moran Mizrahi, Guy Kaplan, Dan Malkin, Rotem Dror, Dafna Shahaf, and Gabriel
  Stanovsky.
\newblock State of what art? a call for multi-prompt {LLM} evaluation.
\newblock \emph{Transactions of the Association for Computational Linguistics},
  12:\penalty0 933--949, 2024.
\newblock \doi{10.1162/tacl_a_00681}.

\bibitem[Muennighoff et~al.(2025)Muennighoff, Yang, Shi, Li, Fei-Fei,
  Hajishirzi, Zettlemoyer, Liang, Cand{\`e}s, and Hashimoto]{muennighoff2025s1}
Niklas Muennighoff, Zitong Yang, Weijia Shi, Xiang~Lisa Li, Li~Fei-Fei,
  Hannaneh Hajishirzi, Luke Zettlemoyer, Percy Liang, Emmanuel Cand{\`e}s, and
  Tatsunori Hashimoto.
\newblock s1: Simple test-time scaling.
\newblock \emph{arXiv preprint arXiv:2501.19393}, 2025.

\bibitem[{OpenAI}(2025{\natexlab{a}})]{gpt5card}
{OpenAI}.
\newblock {GPT-5} system card.
\newblock \url{https://openai.com/index/gpt-5-system-card/},
  2025{\natexlab{a}}.
\newblock Also available as arXiv:2601.03267.

\bibitem[{OpenAI}(2025{\natexlab{b}})]{gptoss2025}
{OpenAI}.
\newblock {gpt-oss-120b} \& {gpt-oss-20b} model card, 2025{\natexlab{b}}.
\newblock URL \url{https://arxiv.org/abs/2508.10925}.

\bibitem[{OpenAI}(2026)]{gpt54card}
{OpenAI}.
\newblock {GPT-5.4} thinking system card.
\newblock \url{https://openai.com/index/gpt-5-4-thinking-system-card/}, March
  2026.

\bibitem[Pezeshkpour and Hruschka(2024)]{pezeshkpour2024large}
Pouya Pezeshkpour and Estevam Hruschka.
\newblock Large language models sensitivity to the order of options in
  multiple-choice questions.
\newblock In \emph{Findings of the Association for Computational Linguistics:
  NAACL 2024}, pages 2006--2017, Mexico City, Mexico, 2024. Association for
  Computational Linguistics.
\newblock \doi{10.18653/v1/2024.findings-naacl.130}.

\bibitem[Press et~al.(2023)Press, Zhang, Min, Schmidt, Smith, and
  Lewis]{press2023measuring}
Ofir Press, Muru Zhang, Sewon Min, Ludwig Schmidt, Noah~A. Smith, and Mike
  Lewis.
\newblock Measuring and narrowing the compositionality gap in language models.
\newblock In \emph{Findings of the Association for Computational Linguistics:
  EMNLP 2023}, pages 5687--5711, Singapore, 2023. Association for Computational
  Linguistics.
\newblock \doi{10.18653/v1/2023.findings-emnlp.378}.

\bibitem[{Qwen Team}(2025)]{qwen3}
{Qwen Team}.
\newblock {Qwen3} technical report.
\newblock \emph{arXiv preprint arXiv:2505.09388}, 2025.

\bibitem[Rein et~al.(2024)Rein, Hou, Stickland, Petty, Pang, Dirani, Michael,
  and Bowman]{rein2023gpqa}
David Rein, Betty~Li Hou, Asa~Cooper Stickland, Jackson Petty, Richard~Yuanzhe
  Pang, Julien Dirani, Julian Michael, and Samuel~R. Bowman.
\newblock {GPQA}: A graduate-level google-proof q\&a benchmark.
\newblock In \emph{First Conference on Language Modeling (COLM)}, 2024.
\newblock arXiv:2311.12022.

\bibitem[Robinson et~al.(2023)Robinson, Rytting, and
  Wingate]{robinson2023leveraging}
Joshua Robinson, Christopher~Michael Rytting, and David Wingate.
\newblock Leveraging large language models for multiple choice question
  answering.
\newblock In \emph{The Eleventh International Conference on Learning
  Representations (ICLR)}, 2023.

\bibitem[Schulhoff et~al.(2024)Schulhoff, Ilie, Balepur, Kahadze, Liu, Si, Li,
  Gupta, Han, Schulhoff, Dulepet, Vidyadhara, Ki, Agrawal, Pham, Kroiz, Li,
  Tao, Srivastava, Da~Costa, Gupta, Rogers, Goncearenco, Sarli, Galynker,
  Peskoff, Carpuat, White, Anadkat, Hoyle, and Resnik]{schulhoff2024prompt}
Sander Schulhoff, Michael Ilie, Nishant Balepur, Konstantine Kahadze, Amanda
  Liu, Chenglei Si, Yinheng Li, Aayush Gupta, HyoJung Han, Sevien Schulhoff,
  Pranav~Sandeep Dulepet, Saurav Vidyadhara, Dayeon Ki, Sweta Agrawal, Chau
  Pham, Gerson Kroiz, Feileen Li, Hudson Tao, Ashay Srivastava, Hevander
  Da~Costa, Saloni Gupta, Megan~L. Rogers, Inna Goncearenco, Giuseppe Sarli,
  Igor Galynker, Denis Peskoff, Marine Carpuat, Jules White, Shyamal Anadkat,
  Alexander Hoyle, and Philip Resnik.
\newblock The prompt report: A systematic survey of prompting techniques, 2024.

\bibitem[Sclar et~al.(2024)Sclar, Choi, Tsvetkov, and
  Suhr]{sclar2024quantifying}
Melanie Sclar, Yejin Choi, Yulia Tsvetkov, and Alane Suhr.
\newblock Quantifying language models' sensitivity to spurious features in
  prompt design or: How i learned to start worrying about prompt formatting.
\newblock In \emph{The Twelfth International Conference on Learning
  Representations (ICLR)}, 2024.

\bibitem[Snell et~al.(2025)Snell, Lee, Xu, and Kumar]{snell2024scaling}
Charlie Snell, Jaehoon Lee, Kelvin Xu, and Aviral Kumar.
\newblock Scaling llm test-time compute optimally can be more effective than
  scaling model parameters.
\newblock In \emph{The Thirteenth International Conference on Learning
  Representations (ICLR)}, 2025.
\newblock arXiv:2408.03314.

\bibitem[Sprague et~al.(2024)Sprague, Yin, Rodriguez, Jiang, Wadhwa, Singhal,
  Zhao, Ye, Mahowald, and Durrett]{sprague2024cot}
Zayne Sprague, Fangcong Yin, Juan~Diego Rodriguez, Dongwei Jiang, Manya Wadhwa,
  Prasann Singhal, Xinyu Zhao, Xi~Ye, Kyle Mahowald, and Greg Durrett.
\newblock To {CoT} or not to {CoT}? chain-of-thought helps mainly on math and
  symbolic reasoning.
\newblock \emph{arXiv preprint arXiv:2409.12183}, 2024.

\bibitem[Srivastava et~al.(2022)Srivastava, Rastogi, Rastogi, Awal Md~Shoeb,
  Abid, Fisch, Brown, Santoro, Gupta, Garriga-Alonso,
  et~al.]{srivastava2022beyond}
Aarohi Srivastava, Abhinav Rastogi, Abhishek Rastogi, Abu Awal Md~Shoeb,
  Abubakar Abid, Adam Fisch, Adam~R. Brown, Adam Santoro, Aditya Gupta,
  Adri{\`a} Garriga-Alonso, et~al.
\newblock Beyond the imitation game: Quantifying and extrapolating the
  capabilities of language models.
\newblock \emph{arXiv preprint arXiv:2206.04615}, 2022.

\bibitem[Stechly et~al.(2025)Stechly, Valmeekam, and
  Kambhampati]{stechly2024self}
Kaya Stechly, Karthik Valmeekam, and Subbarao Kambhampati.
\newblock On the self-verification limitations of large language models on
  reasoning and planning tasks.
\newblock In \emph{The Thirteenth International Conference on Learning
  Representations (ICLR)}, 2025.

\bibitem[Tie et~al.(2025)Tie, Yuan, Zhao, Hu, Gu, Zhang, Zhang, Wu, Tu, Jin,
  Wen, Chen, Zhou, and Sun]{tie2025correctbench}
Guiyao Tie, Zenghui Yuan, Zeli Zhao, Chaoran Hu, Tianhe Gu, Ruihang Zhang,
  Sizhe Zhang, Junran Wu, Xiaoyue Tu, Ming Jin, Qingsong Wen, Lixing Chen, Pan
  Zhou, and Lichao Sun.
\newblock Correctbench: A benchmark of self-correction in llms.
\newblock In \emph{Proceedings of the NeurIPS 2025 Datasets and Benchmarks
  Track}, 2025.

\bibitem[Wang et~al.(2023{\natexlab{a}})Wang, Xu, Lan, Hu, Lan, Lee, and
  Lim]{wang2023planandsolve}
Lei Wang, Wanyu Xu, Yihuai Lan, Zhiqiang Hu, Yunshi Lan, Roy Ka-Wei Lee, and
  Ee-Peng Lim.
\newblock Plan-and-solve prompting: Improving zero-shot chain-of-thought
  reasoning by large language models.
\newblock In \emph{Proceedings of the 61st Annual Meeting of the Association
  for Computational Linguistics (Volume 1: Long Papers)}, pages 2609--2634,
  Toronto, Canada, 2023{\natexlab{a}}. Association for Computational
  Linguistics.
\newblock \doi{10.18653/v1/2023.acl-long.147}.

\bibitem[Wang et~al.(2023{\natexlab{b}})Wang, Wei, Schuurmans, Le, Chi, Narang,
  Chowdhery, and Zhou]{wang2023selfconsistencyimproveschainthought}
Xuezhi Wang, Jason Wei, Dale Schuurmans, Quoc Le, Ed~Chi, Sharan Narang,
  Aakanksha Chowdhery, and Denny Zhou.
\newblock Self-consistency improves chain of thought reasoning in language
  models, 2023{\natexlab{b}}.
\newblock URL \url{https://arxiv.org/abs/2203.11171}.

\bibitem[Wang et~al.(2024)Wang, Ma, Zhang, Ni, Chandra, Guo, Ren, Arulraj, He,
  Jiang, et~al.]{wang2024mmlupro}
Yubo Wang, Xueguang Ma, Ge~Zhang, Yuansheng Ni, Abhranil Chandra, Shiguang Guo,
  Weiming Ren, Aaran Arulraj, Xuan He, Ziyan Jiang, et~al.
\newblock Mmlu-pro: A more robust and challenging multi-task language
  understanding benchmark.
\newblock \emph{arXiv preprint arXiv:2406.01574}, 2024.

\bibitem[Wei et~al.(2022)Wei, Wang, Schuurmans, Bosma, Ichter, Xia, Chi, Le,
  and Zhou]{wei2022chain}
Jason Wei, Xuezhi Wang, Dale Schuurmans, Maarten Bosma, Brian Ichter, Fei Xia,
  Ed~Chi, Quoc Le, and Denny Zhou.
\newblock Chain-of-thought prompting elicits reasoning in large language
  models.
\newblock In \emph{Advances in Neural Information Processing Systems},
  volume~35, pages 24824--24837, 2022.
\newblock URL
  \url{https://proceedings.neurips.cc/paper/2022/hash/9d5609613524ecf4f15af0f7b31abca4-Abstract-Conference.html}.

\bibitem[Yao et~al.(2023)Yao, Yu, Zhao, Shafran, Griffiths, Cao, and
  Narasimhan]{yao2023tree}
Shunyu Yao, Dian Yu, Jeffrey Zhao, Izhak Shafran, Thomas~L. Griffiths, Yuan
  Cao, and Karthik Narasimhan.
\newblock Tree of thoughts: Deliberate problem solving with large language
  models.
\newblock In \emph{Advances in Neural Information Processing Systems},
  volume~36, 2023.

\bibitem[Yasunaga et~al.(2024)Yasunaga, Chen, Li, Pasupat, Leskovec, Liang,
  Chi, and Zhou]{yasunaga2024large}
Michihiro Yasunaga, Xinyun Chen, Yujia Li, Panupong Pasupat, Jure Leskovec,
  Percy Liang, Ed~H. Chi, and Denny Zhou.
\newblock Large language models as analogical reasoners.
\newblock In \emph{The Twelfth International Conference on Learning
  Representations (ICLR)}, 2024.

\bibitem[Yu et~al.(2022)Yu, Iter, Wang, Xu, Ju, Sanyal, Zhu, Zeng, and
  Jiang]{yu2022generate}
Wenhao Yu, Dan Iter, Shuohang Wang, Yichong Xu, Mingxuan Ju, Soumya Sanyal,
  Chenguang Zhu, Michael Zeng, and Meng Jiang.
\newblock Generate rather than retrieve: Large language models are strong
  context generators.
\newblock \emph{arXiv preprint arXiv:2209.10063}, 2022.
\newblock URL \url{https://arxiv.org/abs/2209.10063}.

\bibitem[Zheng et~al.(2024)Zheng, Zhou, Meng, Zhou, and Huang]{zheng2024large}
Chujie Zheng, Hao Zhou, Fandong Meng, Jie Zhou, and Minlie Huang.
\newblock Large language models are not robust multiple choice selectors.
\newblock In \emph{The Twelfth International Conference on Learning
  Representations (ICLR)}, 2024.

\bibitem[Zheng et~al.(2025)Zheng, Cheng, Yao, Wu, He, Ding, Cheng, Hu, Bai,
  Zhou, Cui, and Ye]{physicsqa2023}
Shenghe Zheng, Qianjia Cheng, Junchi Yao, Mengsong Wu, Haonan He, Ning Ding,
  Yu~Cheng, Shuyue Hu, Lei Bai, Dongzhan Zhou, Ganqu Cui, and Peng Ye.
\newblock Scaling physical reasoning with the physics dataset, 2025.
\newblock URL \url{https://arxiv.org/abs/2506.00022}.

\bibitem[Zhu et~al.(2024)Zhu, Zhao, Chen, Wang, and Xie]{zhu2024promptbench}
Kaijie Zhu, Qinlin Zhao, Hao Chen, Jindong Wang, and Xing Xie.
\newblock {PromptBench}: A unified library for evaluation of large language
  models.
\newblock \emph{Journal of Machine Learning Research}, 2024.
\newblock arXiv:2312.07910.

\bibitem[{Zitnik Lab}(2025)]{curebench2024}
{Zitnik Lab}.
\newblock {CURE-Bench}: Competition on reasoning models for drug
  decision-making in precision therapeutics.
\newblock NeurIPS 2025 Competition Track, 2025.
\newblock URL \url{https://curebench.ai}.
\newblock Hugging Face dataset:
  \url{https://huggingface.co/datasets/update0909/cure-bench-reasoning-traces}.

\end{thebibliography}

\appendix

\section{The Self-Analogical template and answer extraction}
\vspace{-3mm}
\label{app:selfanalog}
Self-Analogical is the technique for which answer extraction matters most, and the reason
is visible in its template. We reproduce it verbatim, as used for all 10 datasets:

\begin{quote}\small\ttfamily
Your task is to tackle multiple choice question-answering problems. When presented with a
question and various multiple choice options, recall relevant multi-choice
question-answering problems as examples. Afterward, proceed to solve the initial problem.

\# Problem: \{question\}

\# Instructions:

\#\# Relevant Problems: Recall three examples of question-answering problems that are
relevant to the initial problem. Your problems should be distinct from each other and from
the initial problem (\emph{e.g., involving different failure modes, sensor but same
asset}). For each problem:
\begin{itemize}\setlength\itemsep{0pt}
\item After ``Question: '', describe the problem with multiple choices
\item After ``Answer: '', explain the solution and enclose the answer choice. Your output
must strictly follow this format: \{``answer'': <the list of selected options, e.g.,
[``A'', ``B'', ``C'', ``D'']>\}
\end{itemize}

\#\# Solve the Initial Problem: Question: Copy and paste the initial problem here. Answer:
Explain the solution and enclose the ultimate answer with choice. Your output must strictly
follow this format: \{``answer'': <the list of selected options, e.g., [``A'', ``B'', ``C'',
``D'']>\}
\end{quote}

\textbf{The template requests four answer objects, not one.} It instructs the model to
emit \texttt{\{"answer": ...\}} for each of three practice exemplars and again for the
final answer, all in the same format. A response that follows the instruction therefore
contains four objects in the requested syntax, of which only the last is the model's
answer to the question that was asked. On our Group~A cells, 95.40\% of Self-Analogical
responses contain more than one such object (per-cell range 72.0\% to 100.0\%), and in
72.25\% the first object differs from the last. Reading the first object rather than the
last scores an exemplar in place of the answer and reduces Self-Analogical's measured
accuracy from 52.96\% to 22.50\%, a difference of 30.46~pp. Across all other techniques the
multi-object rate is 1.34\% and the same choice moves accuracy by 0.10~pp on average. This
is why the extraction procedure of \S\ref{sec:metrics} takes the last object: it is
equivalent for every other technique and correct for this one.

\textbf{Domain wording carried from the original.} Our template follows the public release
of Yasunaga et al.~\cite{yasunaga2024large}. Its parenthetical guidance on how exemplars
should differ, ``\emph{involving different failure modes, sensor but same asset}'', is
phrased for industrial sensor diagnostics, and we applied it unchanged to all 10 datasets,
including medicine, mathematics and graduate-level science. The instruction is therefore
well matched to FailureSensorIQ and poorly matched elsewhere. We kept the template fixed
rather than rewriting it per domain, so that every technique is evaluated as published, but
Self-Analogical's numbers should be read as a lower bound on what a domain-adapted template
would achieve. Notably, even under this handicap it is the strongest technique on the
Group~A matched set (Table~\ref{tab:techniques}).

\section{Model configurations}
\vspace{-3mm}
\label{app:models}

\begin{table*}[h]
\centering
\small
\setlength{\tabcolsep}{6pt}
\renewcommand{\arraystretch}{1.1}
\caption{The 27 model configurations evaluated, where a configuration is a model together
with one reasoning setting. \textbf{Group} records how the configuration is used, i.e.,
Group~A are the 8 configurations run with no reasoning requested and Group~B the 4
open-weight configurations run with reasoning enabled, both of which carry all 8 prompting
techniques (\S\ref{sec:matched}); the remaining 15 are evaluated under baseline prompting.
\textbf{Datasets} gives the number of the 10 datasets on which the configuration was
evaluated.}
\label{tab:models}
\begin{tabularx}{\textwidth}{@{}l l l c c@{}}
\toprule
\textbf{Family} & \textbf{Model} & \textbf{Reasoning settings evaluated} & \textbf{Group} & \textbf{Datasets} \\
\midrule
GPT-5    & gpt-5-2025-08-07                & nothink, low, medium & baseline & 10 \\
GPT-5    & gpt-5-4 (preview)               & nothink, low, medium & baseline & 10 \\
Claude   & Claude Sonnet~4                 & nothink, low, medium & baseline & 9 \\
Claude   & Claude Sonnet~4.6               & low, medium, high    & baseline & 2, 2, 6 \\
Gemini   & Gemini~2.5 Pro                  & low, medium          & baseline & 9 \\
Llama    & Llama-3.1-405B                  & nothink              & baseline & 9 \\
\midrule
GPT-OSS  & gpt-oss-20b                     & medium               & B & 10 \\
GPT-OSS  & gpt-oss-120b                    & medium               & B & 10 \\
Qwen3    & Qwen3-8B                        & medium               & B & 10 \\
Qwen3    & Qwen3-30B-A3B-Thinking-2507     & medium               & B & 10 \\
\midrule
Llama    & Llama-3.2-11B-Vision            & nothink              & A & 10 \\
Llama    & Llama-3.3-70B                   & nothink              & A & 10 \\
Llama    & Llama-4-Maverick-17B-128e (fp8) & nothink              & A & 10 \\
Mistral  & Mistral-Large-2512              & nothink              & A & 10 \\
Mistral  & Mistral-Medium-2505             & nothink              & A & 10 \\
Mistral  & Mistral-Small-3.1-24B-2503      & nothink              & A & 10 \\
Granite  & Granite-3.3-8B                  & nothink              & A & 10 \\
Granite  & Granite-4-h-small               & nothink              & A & 10 \\
\bottomrule
\end{tabularx}
\end{table*}

The 15 baseline configurations additionally carry partial Prompt-Repeat coverage, in most
cases on CureBench alone and for Llama-3.1-405B on all 9 of its datasets. That coverage is
too sparse and too unevenly distributed to support a matched comparison, and as such it is
excluded from every comparison reported in this paper.

\section{Implementation details}
\vspace{-3mm}
\label{app:implementation}

All experiments use identical system prompts and decoding parameters across techniques. We use \texttt{temperature=0} and a single sample per (technique, model, dataset, question) cell; we acknowledge that temperature-zero decoding is not strictly deterministic for vendor-hosted reasoning models, but we hold all controllable parameters constant and rely on the matched comparison and large per-technique cell counts to absorb residual non-determinism. Answer extraction uses one shared parser across all techniques; the same regex and fallback logic apply identically to Baseline and to every technique's final-answer field.
\section{Dataset statistics}
\vspace{-3mm}
\label{app:datasets}
\begin{table*}[h!]
\centering
\small
\setlength{\tabcolsep}{4pt}
\renewcommand{\arraystretch}{1.15}
\caption{Dataset statistics. ``Avg.\ opt.'' is the mean number of options per question, with min--max in parentheses.}
\label{tab:dataset_stats}
\begin{tabularx}{\textwidth}{@{}l r c c l Y @{}}
\toprule
\textbf{Name} & \textbf{Rec.} & \makecell{\textbf{Avg.\ Q}\\\textbf{len}} & \makecell{\textbf{Avg.\ opt.}\\\textbf{(min--max)}} & \textbf{Domain} & \textbf{Answer distribution} \\
\midrule
FailureSensorIQ~\cite{failuresensoriq} & 500 & 130 & 4.2 (2--5) & IoT & A:26.4\% B:25.2\% C:19.8\% D:16.2\% E:12.4\% \\
CURE-Bench~\cite{curebench2024} & 779 & 260 & 4.4 (4--5) & Medical & A:23.7\% B:24.4\% C:24.6\% D:20.5\% E:6.7\% \\
CorrectBench~\cite{tie2025correctbench} & 494 & 279 & 4.9 (4--40) & Multiple & A:16.0\% B:19.8\% C:14.6\% D:15.6\% E:10.3\% \\
PhysicsQA~\cite{physicsqa2023} & 45 & 225 & 4.1 (3--6) & Physics & A:13.3\% B:35.6\% C:33.3\% D:17.8\% \\
AIME~2025~QA~\cite{aime2025qa} & 30 & 527 & 4.3 (3--6) & Mathematics & A:40.0\% B:20.0\% C:10.0\% D:10.0\% E:20.0\% \\
MedQA~\cite{jin2021medqa} & 500 & 753 & 5.0 (5--5) & Medical & A:22.0\% B:20.6\% C:23.2\% D:18.8\% E:15.4\% \\
MMLU-Pro~\cite{wang2024mmlupro} & 500 & 262 & 9.6 (3--10) & Multiple & A--J $\in$ [8.6\%, 12.0\%] \\
OpenBookQA~\cite{OpenBookQA2018} & 500 & 55 & 4.0 (4--4) & Multiple & A:27.6\% B:25.2\% C:26.4\% D:20.8\% \\
SuperGPQA~\cite{supergpqa2024} & 500 & 324 & 9.7 (4--10) & Graduate-level & A--J $\in$ [8.0\%, 11.8\%] \\
Time-MQA~\cite{kong2025timemqatimeseriesmultitask} & 500 & 211 & 3.4 (2--4) & Multiple & A:26.2\% B:39.8\% C:27.4\% D:6.6\% \\
\bottomrule
\end{tabularx}
\end{table*}

\section{Discussion}
\vspace{-3mm}
\label{app:discussion}
\subsection{Why are the gains small and interchangeable?}
\vspace{-2mm}
The matched comparison of \S\ref{sec:baseline} has a shape that asks for explanation.
Five techniques that differ greatly in complexity, from a single clause of role framing to
a scaffold that generates three worked exemplars before answering, all improve on direct
prompting by between 3.92 and 4.69~pp, and none separates from the others. The two
techniques that do not improve on it, Prompt-Repeat and Self-Generate, are also the two
that do not ask the model to reason. On the configurations that reason natively, none of
the five does anything measurable at all.

The reading most consistent with that pattern is that these techniques share a single
mechanism, i.e., they move the model from answering directly into generating an
intermediate derivation before committing, and once it is in that mode the particular
scaffold contributes little. What varies between CoT, Plan-and-Solve, a role primer and an
analogical scaffold is the instruction's elaborateness, not whether the model ends up
reasoning, and it is the latter that the accuracy responds to. Two observations support
this over the alternatives. Prompt-Repeat lengthens the prompt substantially without asking
for reasoning and returns nothing, which argues against explanations based on prompt length
or additional attention over the question. And the entire effect vanishes on models that
reason natively, which is what a mechanism of this kind predicts, i.e., there is no mode to
switch the model into because it is already in it.

Self-Generate is the exception that fits the account rather than straining it. It is the
only technique among the eight that \emph{replaces} the context the model conditions on,
by generating a background document and answering from it, rather than adding an
instruction to the existing question. It is also the only technique that reduces accuracy
consistently, by 1.09~pp on Group~A and 2.95~pp on Group~B. Where the other techniques risk
nothing by failing to help, a generated context can be wrong, and the model then answers a
question it has partly rewritten.

We are cautious about pushing this further. Our design varies the prompt and holds
everything else fixed, which identifies that the reasoning trigger is what matters but not
why a particular trigger is worth four points rather than two. Distinguishing among
mechanisms would need interventions on the reasoning trace itself rather than on the
instruction that elicits it.

\subsection{Rethinking ``thinking''}
\vspace{-2mm}
Our variant analysis points to a sharply non-linear return on reasoning tokens. On the two
families whose no-reasoning setting is verified to emit none, enabling reasoning adds
+12.74~pp for GPT-5-4 and +2.10~pp for Claude Sonnet~4, whereas an eightfold increase in
the budget within the on-state, from 1{,}024 to 8{,}192 tokens, adds between 0.48 and
2.10~pp across the four families. The framing of ``test-time compute'' as a single dial
that can be turned up for proportional gains is therefore misleading, i.e., there is one
large step at the point where reasoning is enabled, followed by a long shallow plateau.

Part of the explanation is that the budget is not what binds. At \emph{medium} the
configurations consume between 590 and 1{,}611 reasoning tokens per question on average,
which is 7\% to 20\% of the budget they were granted. A cap the models do not approach
cannot be expected to change their behaviour when it is raised, and thus reporting a
budget setting without reporting the tokens actually spent tells a reader very little.

The size of the enable-reasoning step is itself model-dependent, and by a wide margin, i.e.,
the same intervention is worth six times as much on GPT-5-4 as on Claude Sonnet~4. One
interpretation is that a substantial part of what explicit reasoning contributes is already
being done implicitly by some models during direct generation, so that turning it on
exposes the computation rather than adding it. This would matter for evaluation, because
comparing models through their reasoning variants alone would then favour models whose
reasoning is more visible rather than models whose reasoning is better. Our data are
consistent with that reading but do not establish it, since we observe token counts and
final answers rather than the content of the traces.

Two directions look more promising than budget scaling. First, adaptive budgets driven by
question difficulty would avoid fixed allocations whose marginal returns we measure to be
small, and the spread of reasoning consumption we observe across datasets suggests the
headroom for such adaptation is real. Second, the field needs measures of reasoning
\emph{quality} beyond final-answer correctness, since within the on-state token count and
accuracy are only loosely coupled in our results.

%\clearpage
%\newpage
%\input{checklist.tex}
\end{document}